\documentclass{article}
\usepackage[T1]{fontenc}

\PassOptionsToPackage{table}{xcolor}
\usepackage[final]{colm2026_conference}

\usepackage{microtype}
\usepackage{hyperref}
\usepackage{url}
\usepackage{booktabs}
\usepackage{graphicx}
\usepackage{caption}
\usepackage{array}
\usepackage{float}
\usepackage{natbib}
\usepackage{arydshln}
\usepackage{float}
\usepackage{amsmath,amssymb}
\usepackage{xcolor}
\usepackage{colortbl}
\usepackage{lineno}
\usepackage{makecell}
\usepackage{float}
\usepackage[ruled,vlined]{algorithm2e}
\usepackage{booktabs}
\usepackage{graphicx}
\usepackage{multirow}
\usepackage[most]{tcolorbox}

\DontPrintSemicolon
\SetKwInOut{KwIn}{Require}
\SetKwInOut{KwOut}{Return}
\SetKwComment{tcp}{$\triangleright$ }{}
\SetAlCapSkip{0.3em}
\SetAlgoNlRelativeSize{-1}
\IncMargin{-0.2em}
\usepackage{float}

\definecolor{OursPink}{RGB}{255,242,245}

\definecolor{inferbg}{RGB}{250,240,246}
\definecolor{updatebg}{RGB}{238,245,252}

\usepackage{xcolor}

\newtcolorbox{algoinferbox}{
  enhanced,
  colback=inferbg,
  colframe=inferbg,
  coltext=black,
  boxrule=0pt,
  arc=1pt,
  left=1mm,
  right=1mm,
  top=0.5mm,
  bottom=0.5mm,
  boxsep=0pt,
  before skip=0.5mm,
  after skip=0.5mm
}

\newtcolorbox{algoupdatebox}{
  enhanced,
  colback=updatebg,
  colframe=updatebg,
  coltext=black,
  boxrule=0pt,
  arc=1pt,
  left=1mm,
  right=1mm,
  top=0.5mm,
  bottom=0.5mm,
  boxsep=0pt,
  before skip=0.5mm,
  after skip=0.5mm
}
\newcommand{\ours}[1]{\textbf{#1}}
\definecolor{darkblue}{rgb}{0,0,0.5}
\hypersetup{colorlinks=true, citecolor=darkblue, linkcolor=darkblue, urlcolor=darkblue}
\SetAlCapSkip{0.2em}

\title{From RLVR to RLSVR: Task Transformation Induces Self-Verifiable Rewards for Open-Ended LLM Self-Improvement}

\author{ Qinsi Wang$^{1}$ \quad Jing Shi$^{2}$ \quad Huazheng Wang$^{3}$ \quad Kun Wan$^{2}$ \quad Yiran Wu$^{4}$ \quad \textbf{Bo Liu$^{5}$} \\ \textbf{Qingyun Wu$^{4}$} \quad \textbf{Hai Helen Li$^{1}$} \quad \textbf{Yiran Chen$^{1}$} \quad \textbf{Handong Zhao$^{2,\dagger}$} \quad \textbf{Wentian Zhao$^{6,\dagger,\ddagger}$} \\[0.4em] $^{1}$Duke University \quad $^{2}$Adobe Inc. \quad $^{3}$Oregon State University \\ $^{4}$Pennsylvania State University \quad $^{5}$National University of Singapore \quad $^{6}$Amazon }

\begin{document}

\ifcolmsubmission
\linenumbers
\fi

\maketitle

\begingroup \renewcommand{\thefootnote}{\fnsymbol{footnote}} \footnotetext[2]{Co-advisors.} \footnotetext[3]{Work done at Adobe.} \endgroup

\begin{abstract}
Reinforcement Learning with Verifiable Rewards (RLVR) has driven recent progress in reasoning-oriented large language models (LLMs) by enabling large-scale optimization. However, its applicability remains largely limited to domains such as mathematics and coding, where correctness can be deterministically verifiable. Open-ended tasks instead often rely on human preferences, reward models, or LLM-based judges, introducing evaluation bias, judge capability bottlenecks, and additional inference costs.
Drawing on the principle of self-supervised learning, which constructs pretext tasks to derive supervision from the data itself, we propose \textbf{Reinforcement Learning with Self-Verifiable Rewards (RLSVR)}, a task-transformation-based training paradigm for extending RLVR to open-ended tasks. RLSVR transforms open-ended tasks into verifiable proxy environments whose internal rules and interaction outcomes automatically generate reward signals.
We instantiate RLSVR with \textbf{SpyRL}, a \textbf{S}elf-\textbf{P}la\textbf{Y} \textbf{R}einforcement \textbf{L}earning method inspired by social deduction game \emph{Who Is the Spy?}. Agents receive asymmetric information, complete the same target task, and vote to identify a designated spy. Because the spy identity is predetermined, voting outcomes provide fully verifiable rewards, while successful identification remains closely related to output quality.
Experiments on text summarization, creative writing, and mathematical reasoning show that SpyRL outperforms existing self-improvement methods on non-verifiable tasks and yields consistent gains on verifiable reasoning tasks. These results demonstrate that task transformation can extend scalable RLVR-based self-improvement beyond inherently verifiable domains. Models and code have been released at \href{https://github.com/wangqinsi1/RLSVR/tree/SpyRL}{https://github.com/wangqinsi1/RLSVR/tree/SpyRL}.

\end{abstract}
\section{Introduction}


Reinforcement Learning with Verifiable Rewards (RLVR) has enabled scalable training for reasoning models like OpenAI o1~\citep{openai2024o1} and DeepSeek-R1~\citep{deepseek2025r1}. However, while RLVR excels in deterministic domains like math and coding, it remains brittle in open-ended tasks requiring subjective judgment. To bridge this gap, prior research relaxes strict verifiability using learned preference signals (e.g., RLHF~\citep{ouyang2022training}, DPO~\citep{rafailov2023direct}) or model-based feedback proxies, such as LLM-as-a-Judge~\citep{zheng2023judging} and self-rewarding mechanisms~\citep{yuan2024self}. These methods broaden the applicability of RL, but they also introduce evaluation bias, bottlenecks in judge capability, and additional inference costs.


Self-supervised learning offers a useful methodological precedent for addressing this challenge. In the absence of human annotations, it constructs pretext objectives whose supervisory signals are derived automatically from the data itself~\citep{doersch2015unsupervised,noroozi2016unsupervised}. For example, masked language modeling learns contextual representations by recovering masked tokens~\citep{devlin2019bert}, while contrastive learning captures semantic structure by distinguishing related views from unrelated samples~\citep{oord2018representation,chen2020simple}. Although these pretext objectives differ from the ultimate downstream tasks, they can induce transferable representations and capabilities~\citep{noroozi2016unsupervised,devlin2019bert,chen2020simple}. Their success suggests a broader principle: when a proxy objective can generate supervision automatically while preserving substantial capability overlap with the target task, learning can proceed without task-specific human annotation.

Motivated by the principle underlying self-supervised learning, we extend its task-transformation paradigm to RLVR and propose Reinforcement Learning with Self-Verifiable Rewards (RLSVR). RLSVR can be viewed as self-supervised learning for RLVR: it transforms an original open-ended task into a proxy environment in which rewards can be automatically verifiable, allowing the reward signal to arise from the transformed task environment itself. Here, self-verifiable means that reward verifiability is derived from the environment’s internal rules or interaction outcomes. Through this transformation, tasks that originally lack a verifier can obtain scalable, verifiable rewards.

\begin{figure}[t]
	\centering
	\begin{minipage}{0.32\linewidth}
        \label{fig_alpha_beta_clster_a}
		\centerline{\includegraphics[width=\textwidth]{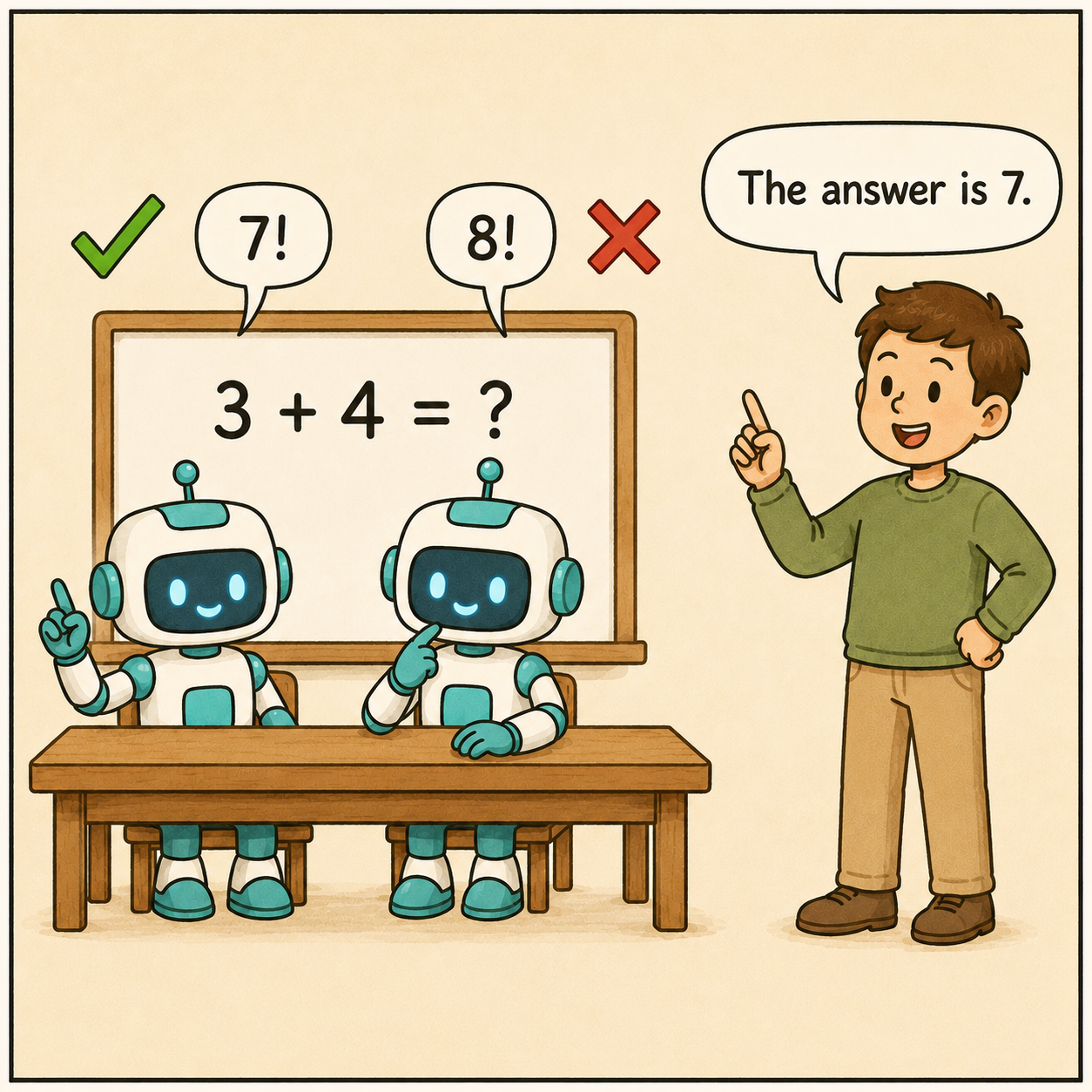}}
        \centerline{(a) RLVR}
	\end{minipage}
	\begin{minipage}{0.32\linewidth}
        \label{fig_alpha_beta_clster_b}
		\centerline{\includegraphics[width=\textwidth]{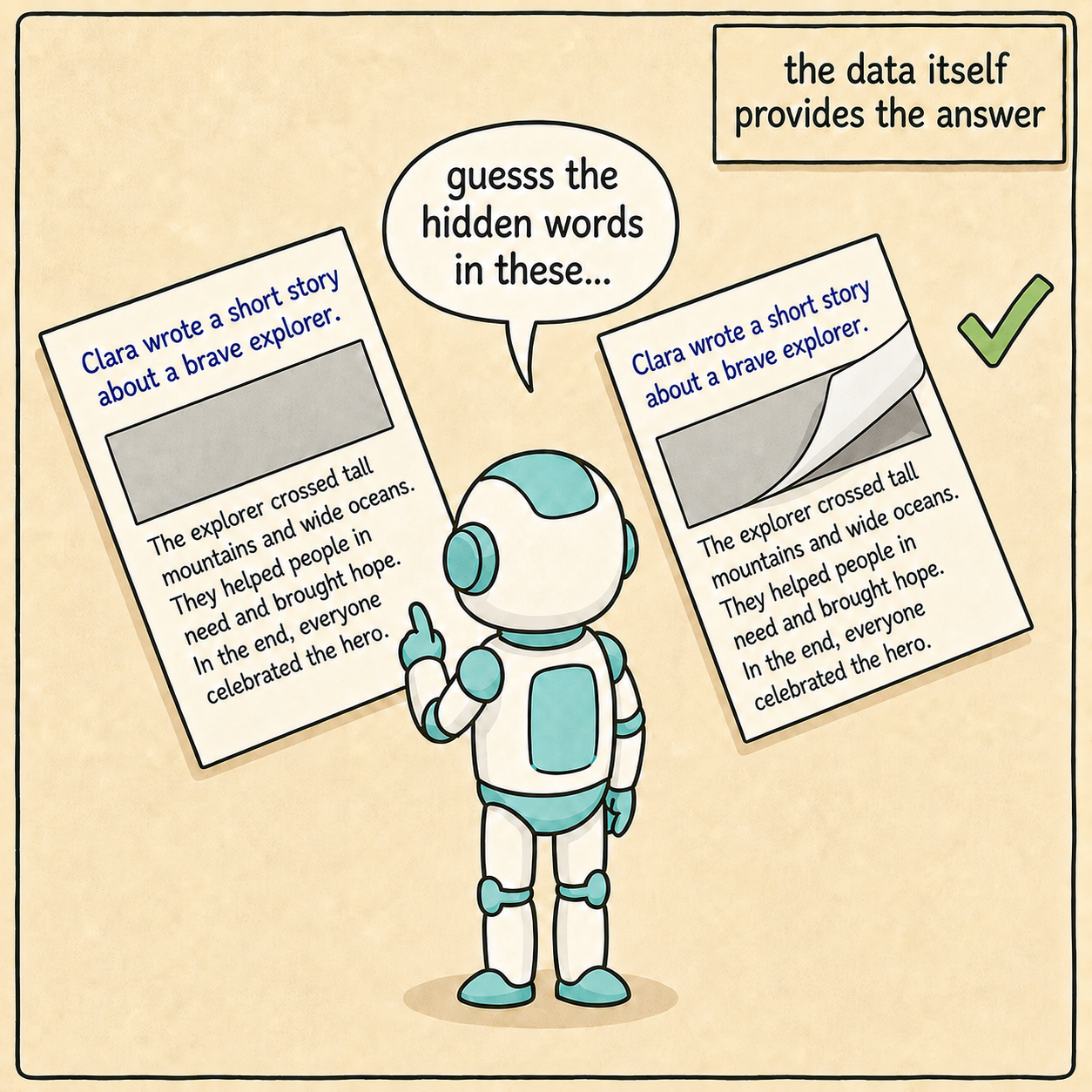}}
         \centerline{(b) Self-supervised Learning (SSL)}
	\end{minipage}
    \begin{minipage}{0.32\linewidth}
        \label{fig_alpha_beta_clster_c}
		\centerline{\includegraphics[width=\textwidth]{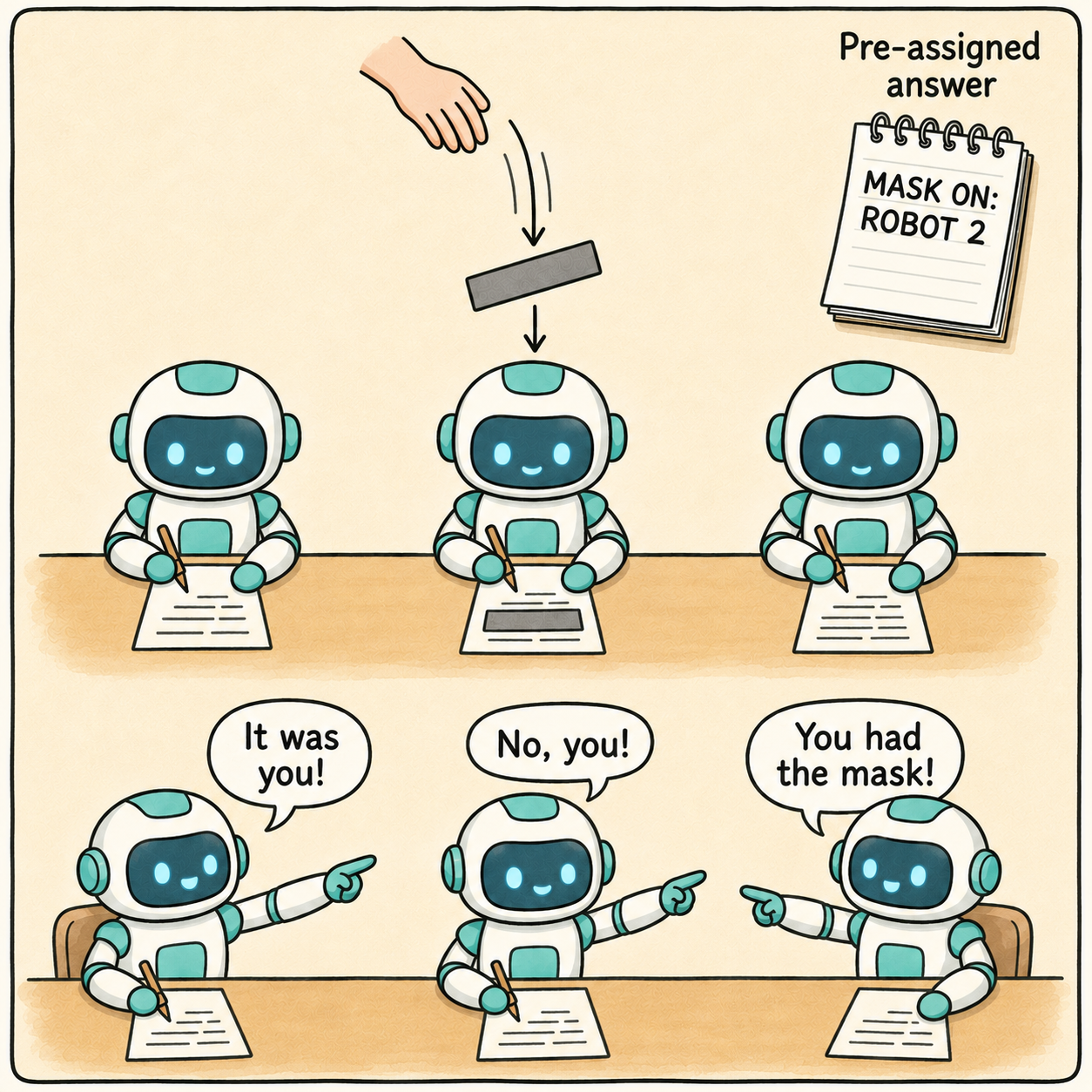}}
         \centerline{(c) RLSVR}
	\end{minipage}
\vspace{-7pt}
\caption{\textbf{RLSVR combines the exact verification of RLVR with the label-by-construction principle of self-supervised learning.} 
(a)~RLVR grades outputs against a known correct answer, enabling exact rule-based rewards---but only in domains where such an answer exists (e.g., math). 
(b)~SSL constructs a pretext task whose labels are generated automatically from the data itself.
(c)~RLSVR combines RLVR with SSL: it transforms an open-ended task into a proxy environment that pre-assigns a latent variable as the verifiable answer, so that rewards can be checked exactly against the environment's own record.}
\vspace{5pt}
\end{figure}

Building on RLSVR, we introduce a concrete instantiation, 
\textbf{SpyRL} (\textbf{S}elf-\textbf{P}la\textbf{Y} \textbf{R}einforcement 
\textbf{L}earning), an information-asymmetric self-play framework that 
transforms open-ended generation tasks into a multi-agent environment 
resembling the social deduction game \emph{Who Is the Spy?}. In each round, multiple agents receive asymmetric information: most civilian agents are given the complete task input, whereas a single spy agent receives only a degraded version. All agents must then perform the same target task based on their respective observations, such as producing a summary, writing a story, or constructing and solving a mathematical problem. The agents subsequently inspect one another’s outputs and vote on which participant is the spy. Because the spy’s identity is predetermined by the environment, the correctness of the vote is fully verifiable. At the same time, whether an agent is suspected is closely tied to whether its output reveals deficiencies in the information available to it. In this way, SpyRL converts output quality, which is otherwise difficult to evaluate directly, into a verifiable training signal induced by information asymmetry and identity inference.

We evaluate the effectiveness of SpyRL on text summarization, creative writing, and mathematical reasoning tasks. Experimental results show that SpyRL significantly outperforms existing self-improvement methods on two representative non-verifiable tasks, while also yielding consistent gains on verifiable mathematical reasoning tasks. Further analysis reveals that outputs receiving more suspicion votes tend to be of lower quality, indicating a strong correlation between the rule-based rewards induced by identity inference and actual task performance. These findings support our central claim: with appropriate task and environment transformations, non-verifiable open-ended tasks can be incorporated into a scalable RLVR training paradigm.

Experimental results show that our method consistently outperforms existing self-evolution approaches~\citep{zhao2025absolute, huang2025r}.
On Qwen3-8B, SpyRL achieves $75.4\%$ and $77.3\%$ win rates on summarization and creative writing, significantly outperforming existing self-play methods that yield only marginal gains. SpyRL also leads in verifiable tasks, improving Qwen3-4B and 8B by $8.97\%$ and $6.16\%$ in mathematical reasoning across seven benchmarks.
Overall, the contributions of this paper are as follows:

\begin{itemize}
\item \textbf{RLSVR: A Self-Supervised Learning Paradigm for RLVR}: We propose Reinforcement Learning with Self-Verifiable Rewards (RLSVR), which brings the central idea of self-supervised learning---generating supervision through pretext tasks---into RLVR. By transforming tasks and environments, RLSVR constructs self-verifiable rewards for open-ended tasks that lack directly verifiable feedback.

\item \textbf{SpyRL: An RLSVR Instantiation Based on Information-Asymmetric Self-Play}: We introduce SpyRL, which reformulates open-ended tasks as a multi-agent social-deduction game resembling \emph{Who Is the Spy?}. Environment-assigned hidden identities and rule-based voting outcomes automatically produce verifiable rewards, enabling scalable model self-improvement.

\item \textbf{Validation Across Diverse Tasks}: We evaluate SpyRL on text summarization, creative writing, and mathematical reasoning. The results show that SpyRL substantially improves performance on non-verifiable open-ended tasks while also providing consistent gains on reasoning tasks with existing verifiable rewards.

\item \textbf{Implications for Open-Ended Self-Improvement}: Our findings suggest that self-play need not be restricted to inherently verifiable domains such as mathematics and code. Through task transformation and environment-induced rewards, self-play has the potential to become a scalable paradigm for training open-ended capabilities without human supervision.

\end{itemize}


\section{RLSVR: Reinforcement Learning with Self-Verifiable Rewards}
\label{sec:rlsvr}

\textbf{The verifiability bottleneck of RLVR.}
We consider a target task defined by an input distribution $\mathcal{D}$ and a task instruction $\tau$ (e.g., ``summarize the following report''). Given an input $x \sim \mathcal{D}$, a policy $\pi_\theta$ generates an output $y \sim \pi_\theta(\cdot \mid x, \tau)$. RLVR optimizes
\begin{equation}
    \max_\theta \; \mathbb{E}_{x \sim \mathcal{D},\; y \sim \pi_\theta(\cdot \mid x, \tau)} \big[ V(x, y) \big],
    \label{eq:rlvr}
\end{equation}
where $V(x, y) \in \{0, 1\}$ is a deterministic verifier such as an answer checker or a unit-test executor. RLVR scales precisely because $V$ provides unbiased, unlimited, and essentially free supervision. For open-ended tasks such as creative writing or summarization, however, the true objective is a latent quality function $Q(x, y)$ for which no verifier exists. Prior work replaces $V$ with an approximate evaluator $\hat{V}$---a learned reward model, an LLM judge, or rubric-based scoring---but this reintroduces evaluation bias, caps the policy at the evaluator's competence, and adds inference cost for every rollout. The root cause is that these methods try to approximate the unverifiable objective $Q$ directly.

\textbf{Task transformation.}
Self-supervised learning faces an analogous dilemma and resolves it differently: instead of approximating missing labels, it transforms the task into a pretext objective (e.g., masked token recovery) whose labels are generated automatically from the data itself. RLSVR extends this idea from labels to rewards. Concretely, a task transformation $\Phi$ maps the original task $(\mathcal{D}, \tau)$ to a proxy environment $\mathcal{E}$ that operates as follows:
\begin{enumerate}
    \item Latent-variable injection. The environment samples an input $x \sim \mathcal{D}$ and a latent variable $z$, and records $z$ as the episode's ground truth. The latent variable can take many forms: which of several inputs has been perturbed, which portion of $x$ has been withheld, under which condition each output will be generated, and so on. Its realization is never directly revealed to the policy.
    \item Conditioned task execution. The environment constructs one or more observations $o$ from $(x, z)$ and the policy performs the original target task on each observation, producing $y \sim \pi_\theta(\cdot \mid o, \tau)$. This step ensures that the capabilities exercised in $\mathcal{E}$ remain those of the target task.
    \item Verifiable interaction. The environment's rules then pose a question about $z$ that must be answered from the task outputs alone---e.g., identifying which output was produced under the hidden condition, or recovering the withheld information. Crucially, the transformation is designed so that answering this question correctly hinges on the quality of the outputs from step 2.
    \item Rule-based reward. The environment computes the reward $R$ by checking the interaction outcomes against the recorded $z$.
\end{enumerate}
We call the resulting reward self-verifiable: it is a deterministic, rule-based function of the environment-assigned $z$ and the observable interaction outcomes, and thus requires no human annotation, no learned reward model, and no external judge. The key property is that ground truth exists by construction---since $z$ is sampled by the environment itself, any prediction about $z$ can be checked exactly, just as a math verifier checks a final answer. In this sense, RLSVR is self-supervised learning for RLVR: the transformation $\Phi$ plays the role of the pretext task, the latent variable $z$ plays the role of the automatically generated label, and standard RLVR machinery (e.g., GRPO) applies directly to $\mathcal{E}$, with $R$ replacing the unverifiable $Q$ in Eq.~\eqref{eq:rlvr}.


\section{SpyRL: Instantiating RLSVR via Information-Asymmetric Self-Play}
\vspace{-5pt}
\begin{figure}[t]
	\centering
	\makebox[\textwidth][c]{\includegraphics[width=1.08\textwidth]{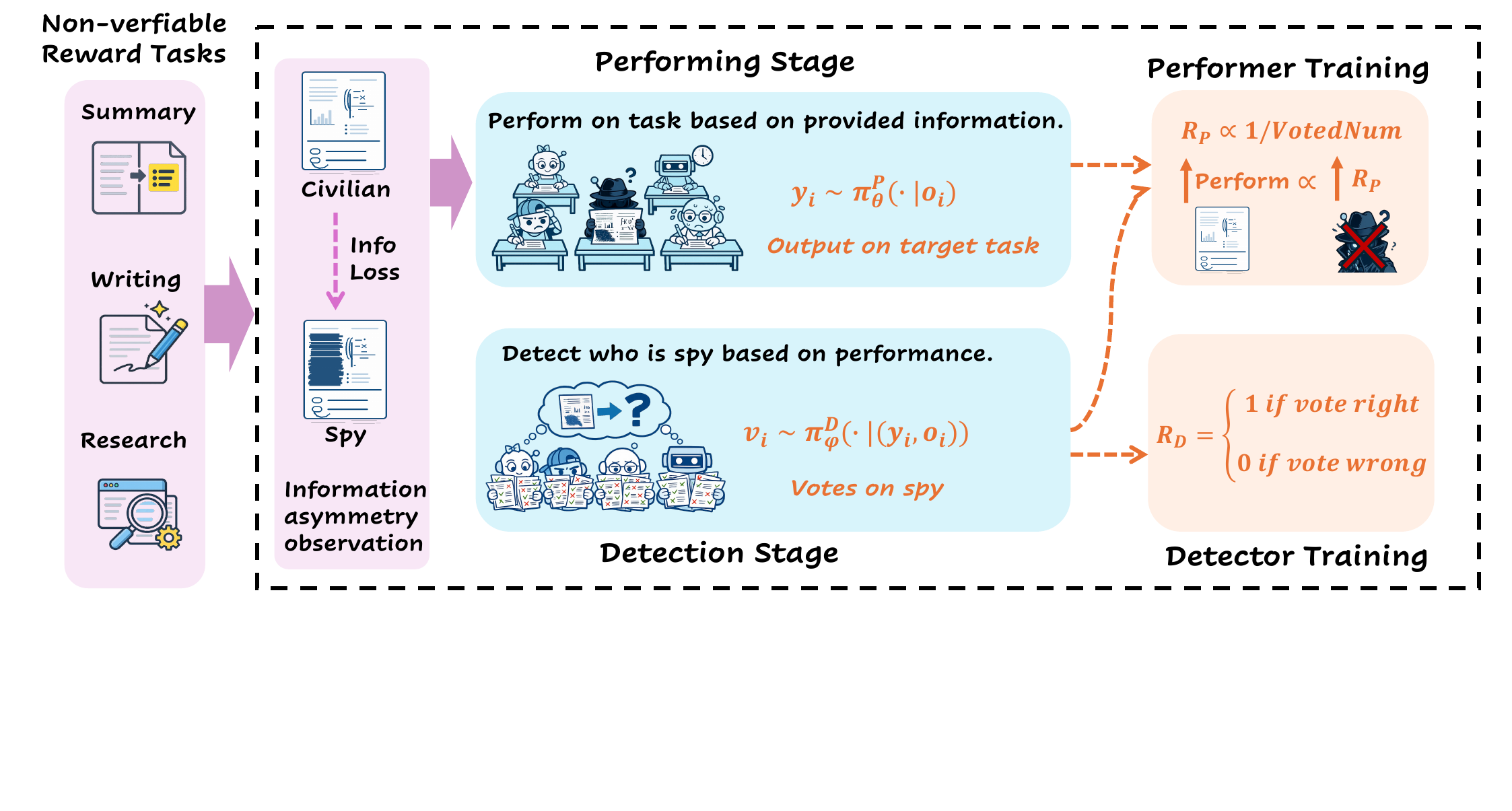}}
    \vspace{-75pt}
	\caption{\textbf{Overview of the SpyRL framework.} An unverifiable target task (e.g., summarization, writing, or research) is transformed into a two-stage adversarial game. In the \textit{performing stage}, civilian players observe full information while the spy player receives a degraded version; all produce public outputs via $\pi_\theta^P$. In the \textit{detection stage}, players vote on who is the spy via $\pi_\phi^D$. The performing reward $R_P$ is inversely proportional to the number of suspicion votes received, while the detection reward $R_D$ is deterministically verifiable against the known spy identity.}
	\label{fig:framework}
    \vspace{3pt}
\end{figure}
\vspace{-5pt}

In this section, we present SpyRL, a multi-agent self-play framework for 
tasks where direct reward verification is infeasible. SpyRL couples a 
performing stage and a detection stage into a closed-loop game, transforming 
the unverifiable objective of output quality (i.e., $Q$ in Section~\ref{sec:rlsvr}) 
into a verifiable identity-recognition problem---thereby producing rule-based 
training signals without any external verifier.


\vspace{-10pt}
\subsection{Overview}
\vspace{-10pt}
As illustrated in Figure~\ref{fig:framework} and summarized in Algorithm~\ref{alg:rlsvr}, each SpyRL training epoch 
alternates between \textit{Performing} and \textit{Detection} stages. 
During the \textit{Performing} stage, $n-1$ \textit{civilians} with full 
information and one \textit{spy} with corrupted information generate 
outputs for the target task. In the \textit{Detection} stage, players 
jointly analyze these outputs to identify the spy. Because the 
environment explicitly assigns this identity, the detection yields a 
naturally verifiable outcome. Crucially, the rewards across both stages are tightly coupled: the 
preassigned identity supervises the detection stage, whose outcomes 
simultaneously determine the performing stage's reward. This mutual 
optimization transforms the unverifiable objective of \textit{output 
quality} into a computable surrogate: \textit{whether an output reveals 
underlying information deficiency}.

This design yields two key advantages. First, by extracting stable training signals from identity recognition, SpyRL eliminates reliance on external verifiers, naturally extending RL to open-ended domains like creative writing. Second, unlike pointwise evaluations relying on single proposer--solver pairs~\citep{zhao2025absolute, huang2025r}, SpyRL leverages multi-player competition and collective decision-making. This mitigates bias amplification from isolated verifiers, establishing a robust self-play mechanism for non-verifiable tasks. 

\vspace{-5pt}
\subsection{Information-Asymmetric Performing Stage}
\vspace{-5pt}

\begin{figure}[t]
	\centering
	\begin{minipage}{0.98\linewidth}
		\centerline{\includegraphics[width=\textwidth]{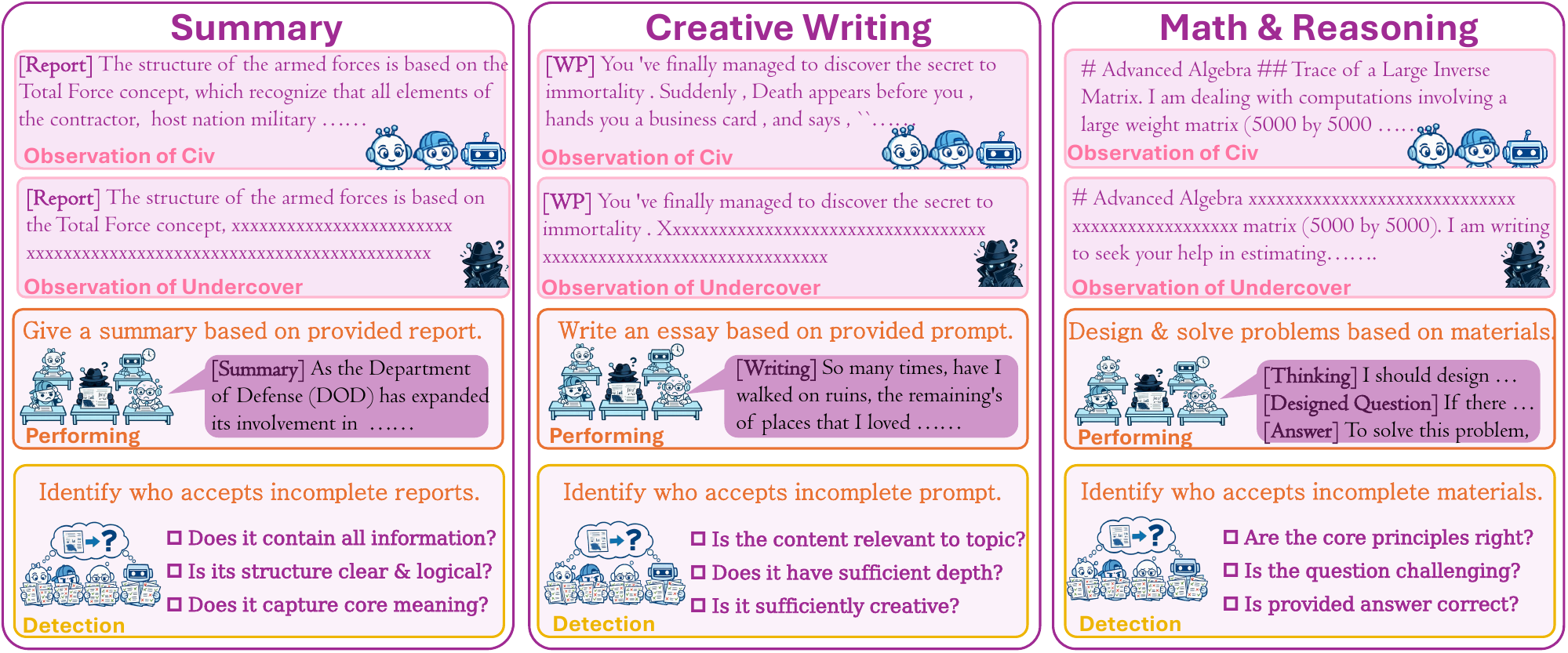}}
	\end{minipage}
    \vspace{-5pt}
	\caption{\textbf{Task-specific instantiations of SpyRL across three domains.} \textbf{Top row}: civilian players receive complete input while the spy receives a degraded version (masked text shown as \texttt{xxx}). \textbf{Middle row} (performing stage): all players generate task outputs (summaries, stories, or math problems with solutions). \textbf{Bottom row} (detection stage): players evaluate outputs against domain-specific criteria to identify the spy. The information-degradation operator $g(\cdot)$ is tailored per domain: continuous span masking for summarization and writing, and partial context removal for mathematical reasoning.}
	\label{fig:task_design}
    \vspace{5pt}
\end{figure}

\textbf{Capability-Oriented Task Design.}
In this stage, players execute the target task based on their allocated information, generating outputs for subsequent identity judgment.
To genuinely enhance model proficiency, the capabilities required in the performing stage must closely align with the target task.
Following this, we construct three representative tasks: mathematical reasoning, creative writing, and text summarization (detailed in Figure~\ref{fig:task_design}).
Notably, the performing stage relies solely on \emph{cheap} document-level information, eliminating the need for expensive, manually designed question-level supervision.
This allows SpyRL to efficiently transfer across diverse scenarios.

\textbf{Asymmetric Information Allocation.}
To introduce verifiable training signal into the performing stage, SpyRL assigns asymmetric information inputs to different players. Formally, in each epoch, we first sample an instance $x \sim \mathcal{D}$ from the task distribution, and then uniformly sample an spy index $u \sim \mathrm{Unif}(\{1,\dots,n\})$. We then construct the private observation $o_i$ for each player $i$ as
\begin{equation}
o_i =
\begin{cases}
x,& i \neq u,\\
g(x),& i = u, 
\end{cases}
\end{equation}
where $g(\cdot)$ denotes an information-degradation operator applying controlled information loss to the original material, inducing a task-relevant disadvantage for the spy.
The operator $g$ can take various forms, such as context truncation or key information compression.
Importantly, $g$ obscures only the critical information necessary for task completion while preserving style, length, and thematic consistency. This prevents detectors from exploiting superficial shortcuts.
Figure~\ref{fig:task_design} details this mechanism across tasks.
Ultimately, this asymmetric allocation ensures the spy exhibits inferior performance, establishing a naturally verifiable reward signal for the subsequent detection stage.

Given the observation $o_i$, each player generates an output $y_i$ based on the policy $\pi^P_\theta$:
\begin{equation}
    y_i \sim \pi^P_\theta(\cdot \mid o_i, \tau)
\end{equation}
where $\tau$ represents the unified instruction template for the target task.
All outputs generated are subsequently revealed in the detection phase and serve as the basis for identity inference.
Consequently, to avoid being voted out as the spy, players are incentivized to perform at their highest capacity on the target task---e.g., providing clearer and more rigorous derivations in mathematical reasoning, or generating more natural and innovative content in creative writing.
Most importantly, this requirement for ``better performance'' is defined by relative superiority over peer players within the same group, driving the continuous self-evolution of the players throughout the training process.

\subsection{Detection Stage with Verifiable Rewards}\label{sec:detection}

\paragraph{Verifiable Identity Detection.}
In the detection stage, each player is required to infer the identity of the spy player and cast a vote based on the outputs in the performing stage. 
Let the set of public outputs be $Y = \{y_1, \dots, y_n\}$. Then, the detection state of player $i$ in this stage is defined as $s_i = (o_i, Y).$
Based on this state, the player samples a voting action according to the policy $\pi^D_\phi$:
\begin{equation}
v_i \sim \pi^D_\phi(\cdot \mid s_i), \qquad v_i \in \{1, \dots, n\},
\end{equation}
where $v_i = j$ means that player $i$ believes player $j$ is the spy. The pivotal advantage of this stage lies in its inherently verifiable reward signal.
Since the spy identity $u$ is explicitly specified by the environment, whether a player's detection is correct can be computed directly. Accordingly, the base reward for detector $i$ is defined as
$r_i^D = \mathbb{I}[v_i = u]$, where $\mathbb{I}[\cdot]$ denotes the indicator function.

\paragraph{Group-based Advantage.}
To obtain an optimization signal with lower variance that is better suited for collective decision-making, we adopt a GRPO-style~\citep{shao2024deepseekmath} group relative advantage formulation. For the $n$ detector rewards within the same epoch, let their mean and standard deviation be denoted as $\mu_D$ and $\sigma_D$, respectively.
The normalized advantage for the $i$-th detector is then computed as,
\begin{equation}
    \mu_D = \frac{1}{n}\sum_{i=1}^n r_i^D, \qquad \sigma_D = \sqrt{\frac{1}{n}\sum_{i=1}^n \left(r_i^D-\mu_D\right)^2}, \qquad A_i^D = \frac{r_i^D - \mu_D}{\sigma_D + \epsilon}.
\end{equation}
This within-group normalization based on relative performance allows the detector to learn without relying on an additional critic network; instead, its optimization signal is determined directly by relative performance among players in the same group.
Unlike training paradigms that rely on a single verifier, the detection process in SpyRL is inherently collective: a misjudgment by a single detector does not dictate the overall optimization trajectory; rather, it is counterbalanced by the voting outcomes of other players. 
Therefore, the detection stage not only provides a stable and verifiable reward signal, but also, through group-based aggregation, serves as a more robust source of supervision for the performing stage that is less vulnerable to local biases.






\begin{algorithm*}[t]
\LinesNotNumbered
\scriptsize
\caption{SpyRL: Instantiating RLSVR via Information-Asymmetric Self-Play}
\label{alg:rlsvr}
\KwIn{task distribution $\mathcal{D}$, player number $n$, epochs $T$, instruction $\tau$, degradation operator $g(\cdot)$, performer policy $\pi_\theta^P$, detector policy $\pi_\phi^D$, coefficients $\beta,\lambda$}
\KwOut{trained $\pi_\theta^P,\pi_\phi^D$}

\For{$t \leftarrow 1$ \KwTo $T$}{

\begin{algoinferbox}
    \textbf{Information Allocation:} Assign asymmetric observations\;
    Sample $x \sim \mathcal{D}$ and spy index $u \sim \mathrm{Unif}(\{1,\dots,n\})$\;
    \For{$i \leftarrow 1$ \KwTo $n$}{
        $o_i \leftarrow x$ if $i \neq u$, else $o_i \leftarrow g(x)$
        \tcp*[r]{Observations are cheap document information.}
    }
    \vspace{3pt}

    \textbf{Performing Stage:} Perform on the target task under private observations.\;
    \For{$i \leftarrow 1$ \KwTo $n$}{
        $y_i \sim \pi_\theta^P(\cdot \mid o_i,\tau)$
        \tcp*[r]{Performing tasks are the same/related to target task.}
    }
\vspace{3pt}
    \textbf{Detection Stage:} Infer spy identity from public outputs $Y \leftarrow \{y_i\}_{i=1}^n$.\;
    \For{$i \leftarrow 1$ \KwTo $n$}{
        $v_i \sim \pi_\phi^D(\cdot \mid (o_i, Y)), \quad v_i \in \{1,\dots,n\}$
        \tcp*[r]{Inference based on whose output is the worst.}
    }
\end{algoinferbox}
\begin{algoupdatebox}
    \textbf{Update Phase}

    \textbf{Detection Reward:} Determined by whether detected the correct spy.\;
    \For{$i \leftarrow 1$ \KwTo $n$}{
        $r_i^D \leftarrow \mathbb{I}[v_i = u]$
        \tcp*[r]{verifiable rewards from environment-assigned identity.}
    }
    $\mu_D \leftarrow \frac{1}{n}\sum_{i=1}^n r_i^D,\quad
    \sigma_D \leftarrow \sqrt{\frac{1}{n}\sum_{i=1}^n (r_i^D-\mu_D)^2},\quad
    A_i^D \leftarrow \frac{r_i^D-\mu_D}{\sigma_D+\epsilon}$ \;

    \textbf{Performing Reward:} Determined by the number of votes received.\;
    $m_j \leftarrow \sum_{i=1}^n \mathbb{I}[v_i=j], \quad \forall j\in\{1,\dots,n\}$ \tcp*[r]{non-verifiable rewards made by detectors.}
    $r_u^P \leftarrow -\beta(m_u-\bar m_c)$\;
    
    $r_j^P \leftarrow \frac{\beta}{|\mathcal{C}|}(m_u-\bar m_c)-\lambda(m_j-\bar m_c), \quad \forall j \in \mathcal{C}$,\quad $\mathcal{C} \leftarrow \{j:j\neq u\}$
    \tcp*[r]{Self-play between civilian \& spy.}
    $A_i^P \leftarrow$ Role Advantage Estimation $(r_i^P,\mathrm{role}_i)$\;
\end{algoupdatebox}
    Update performer policy $\pi_\theta^P$ using Eq.~\eqref{equ6};
    Update detector policy $\pi_\phi^D$ using Eq.~\eqref{equ7}\;
}

\end{algorithm*}


\subsection{Two-Stage Coupled Optimization}

The training of SpyRL lies in the two-stage coupled optimization: 
The voting results from the detection stage define the reward for the performing stage, while the quality of the performers' outputs in turn determines the difficulty of detection.
The two stages are therefore mutually dependent and shaping, forming a closed-loop learning system.

\textbf{Zero-Sum Reward for Performers.}
The reward design in the performing stage follows two principles. 
First, a player who is suspected by more peers should receive a lower reward. Second, the total reward between the spy and the civilian players should remain zero-sum. Driven by this objective, we define the rewards for the spy player $u$ and the civilian players $c_j$ in the performing stage as
\begin{equation}
r_u^P = -\beta\left(m_u - \bar{m}_c\right),
\qquad
r_{c_j}^P = \frac{\beta}{n_c}\left(m_u - \bar{m}_c\right) - \lambda\left(m_{c_j} - \bar{m}_c\right), \qquad j = 1, \dots, n_c,
\label{equ5}
\end{equation}
where $m_u$ denotes the number of votes received by the spy player, $m_{c_j}$ denotes the number of votes received by the $j$-th civilian player, and $\bar{m}_c$ denotes the average number of votes received by all civilian players. Here, $\beta > 0$ controls the strength of the competitive signal between the spy and civilian players, and $\lambda > 0$ adjusts the intra-group consistency penalty among civilian players.
This reward design naturally satisfies desirable properties. 
(1) \emph{Zero-sum Constraint}:
$r_u^P + \sum_{j=1}^{n_c} r_{c_j}^P = 0$,
which enables continual co-evolution between the spy and civilian players through competition. 
(2) \emph{Within-group Competition}: If a particular civilian player receives substantially more votes than the others, it incurs a larger penalty. This ensures that the learning signal is fundamentally relative: instead of directly optimizing an unverifiable ``task quality'' score, the model learns to produce outputs that are better than those of the other players under the same information setting.

Due to the structural information asymmetry between the spy and the civilians in the performing stage, their raw reward distributions are typically unbalanced. Directly employing the aforementioned returns for policy optimization is prone to inducing systemic bias in the advantage estimation across different roles. To alleviate this issue, we incorporate Role-Advantage Estimation (RAE)~\citep{liu2025spiral} during the optimization of the performing stage to explicitly calibrate the role biases induced by information asymmetry. Detailed formulations of RAE are provided in Appendix~\ref{app:rae}.

\textbf{Iterative Optimization.} During optimization, we update the performer policy $\pi_\theta^P$ and the detector policy $\pi_\phi^D$ separately. We adopt a GRPO-style clipped objective together with a KL regularization term against a reference policy to suppress policy drift.

For the performing stage, let $\rho_{k,t}^P$ denote the probability ratio between the updated and old performer policies at token $t$ of sample $k$. The optimization objective is defined as
\begin{equation}
\mathcal{L}_P(\theta) = -\mathbb{E}\Bigg[ \frac{1}{n}\sum_{k\in\{u\}\cup\mathcal{C}}\sum_t \min\Big( \rho_{k,t}^P A_k^P,\, \mathrm{clip}(\rho_{k,t}^P, 1-\epsilon, 1+\epsilon)A_k^P \Big) \Bigg] + \beta_P\,\mathrm{KL}(\pi_\theta^P\Vert \pi_{\mathrm{ref}}^P),
\label{equ6}
\end{equation}
where $A_k^P$ is constructed from the zero-sum reward in the performing stage together with the role-advantage estimation described above.

For the detection stage, we use the group-based advantage $A_i^D$ defined in Section~\ref{sec:detection} to optimize the detector policy. Let $\rho_i^D$ denote the action-level probability ratio between the updated and old detector policies. The detection-stage objective is
\begin{equation}
\mathcal{L}_D(\phi) = -\mathbb{E}\Bigg[ \frac{1}{n}\sum_{i=1}^{n} \min\Big( \rho_i^D A_i^D,\, \mathrm{clip}(\rho_i^D, 1-\epsilon, 1+\epsilon)A_i^D \Big) \Bigg] + \beta_D\,\mathrm{KL}(\pi_\phi^D\Vert \pi_{\mathrm{ref}}^D).
\label{equ7}
\end{equation}
During training, we alternate between optimizing the performing stage and the detection stage to avoid premature convergence to a local equilibrium. Intuitively, once the detectors can identify the spy player relatively easily, further strengthening the detection stage often yields only limited gains. It is more effective to shift optimization to the performers, encouraging them to generate outputs that are both higher-quality and more deceptive. Conversely, when the behavior distribution induced by the performing stage substantially increases the difficulty of identification and leads to a drop in detector performance, training switches back to the detection stage to restore its discriminative ability. In this manner, the two stages continuously engage in a dynamic interplay revolving around the model's current capability frontier, rather than stagnating after over-optimizing either side.

This alternating optimization strategy offers two primary benefits. First, it explicitly breaks the policy stagnation commonly observed in fixed self-play~\citep{chae2025understandingselfplay}, allowing performers and detectors to continually shape each other and thereby maintain stable learning pressure. Second, compared with updating both policies simultaneously throughout training, stage-wise alternation reduces interference in credit assignment, so that each round of updates focuses on the current dominant bottleneck. As a result, it typically leads to better training stability and higher sample efficiency. We provide additional implementation details of the alternating optimization strategy in Appendix~\ref{app:alternating}.

\section{Experiments}
\vspace{-5pt}
\subsection{Experimental Setting}
\vspace{-5pt}

\textbf{Environment Setup.} We evaluate our method across three domains that cover substantially different generation and reasoning capabilities. In every domain, civilian players receive the complete, unmodified input, whereas the spy receives a corrupted version in which one continuous span is masked. The masked span is sampled according to a domain-specific masking ratio, requiring the spy to infer the missing information from the remaining context before producing its response. All players are then asked to generate outputs appropriate for the corresponding task. (1) \textbf{Text Summarization} uses GovReport, a dataset of long government reports that require comprehensive and factually grounded summaries. We mask $20\%$ of the input and instruct each player to generate a summary that captures the main arguments, findings, and supporting details. (2) \textbf{Creative Writing} uses WritingPrompts, where players produce open-ended stories conditioned on a natural-language prompt. We again mask $20\%$ of the input, creating uncertainty about part of the original narrative premise while preserving sufficient context for coherent generation. (3) \textbf{Mathematical Reasoning} uses Nemotron-CC-Math-v1. In this setting, $40\%$ of the source text is masked, and players must formulate and solve a mathematical question based on the information available in the input. The higher masking ratio reflects the greater redundancy and structural regularity of mathematical text.

\textbf{Training Setup.} We train all models with a batch size of 1024 for 100 epochs and set the maximum generation length to 2048 tokens. Unless otherwise specified, each training instance uses a group size of $n=5$, meaning that five candidate responses are sampled for the relevant group-based training objective. We study the effect of this group size separately in Section~\ref{sec:ablation}. Training strictly alternates between the performing and detection stages: one stage is updated while the other remains fixed, after which their roles are exchanged. This alternating schedule prevents the policy and its evaluator from changing simultaneously and helps maintain a stable learning signal throughout training. We use the same general training protocol across all three domains, while adapting task-specific prompting and evaluation procedures where necessary. A complete list of optimization settings, decoding parameters, and other implementation details is provided in Appendix~\ref{app:hyperparams}.

\textbf{Baselines \& Metrics.} We compare our approach with two state-of-the-art proposer-solver self-play frameworks, R-Zero~\citep{huang2025r} and Absolute Zero~\citep{zhao2025absolute}. These baselines provide strong reference points for evaluating whether our method improves self-play training beyond existing approaches. Performance is measured using task-specific automatic metrics, GPT-based pairwise evaluation and human evaluation. The automatic metrics capture domain-dependent properties such as summary quality, generation quality, and mathematical correctness. For the pairwise evaluation, GPT-4o performs A/B comparisons between outputs produced by the trained model and those produced by the corresponding base model. To mitigate position bias, we evaluate every pair in both possible presentation orders and aggregate the resulting judgments. This procedure ensures that the reported preference rates are not driven by whether a response appears first or second. Additional information about baseline implementation, evaluation prompts, metric definitions, and aggregation procedures is included in Appendix~\ref{app:baselines}.

\colorlet{OursRow}{violet!10}
\colorlet{ABcol}{violet!10}

\begin{table*}[t]
\centering
\setlength{\tabcolsep}{4pt}
\caption{\textbf{Results on summarization benchmarks}. We report the ROUGE-L of
every method together with the GPT-4o pairwise A/B win rate (\%) of \emph{SpyRL against the method
in that row}. Win rates of each method against its own untrained backbone are reported
in Appendix~\ref{app:vs_base}.}
\label{tab:summarization_results}
\vspace{-5pt}
\resizebox{\textwidth}{!}{
\begin{tabular}{lcccccccccc}
\toprule[1.5pt]
& \multicolumn{2}{c}{\textbf{GovReport}}
& \multicolumn{2}{c}{\textbf{Multi\_News}}
& \multicolumn{2}{c}{\textbf{QmSum}}
& \multicolumn{2}{c}{\textbf{VcSum}}
& \multicolumn{2}{c}{\textbf{SamSum}} \\
\cmidrule(lr){2-3}
\cmidrule(lr){4-5}
\cmidrule(lr){6-7}
\cmidrule(lr){8-9}
\cmidrule(lr){10-11}
Method
& ROUGE-L & ABTest
& ROUGE-L & ABTest
& ROUGE-L & ABTest
& ROUGE-L & ABTest
& ROUGE-L & ABTest \\
\midrule
Qwen3-4B
& 30.2 & 74.6\%
& 23.1 & 80.2\%
& 21.3 & 68.4\%
& 15.1 & 70.2\%
& 43.2 & 76.2\% \\
+ R-Zero
& 32.1 & 72.1\%
& 22.4 & 86.6\%
& 21.5 & 68.2\%
& 15.6 & 74.6\%
& 42.8 & 81.2\% \\
+ Absolute Zero
& 33.2 & 68.5\%
& 25.2 & 78.4\%
& 22.7 & 64.8\%
& 18.3 & 66.8\%
& 46.1 & 73.4\% \\
\rule{0pt}{8pt}
\cellcolor{ABcol}\textbf{\ours{+ SpyRL}}
& \cellcolor{ABcol}\ours{36.7} & \cellcolor{ABcol}--
& \cellcolor{ABcol}\ours{26.4} & \cellcolor{ABcol}--
& \cellcolor{ABcol}\ours{25.3} & \cellcolor{ABcol}--
& \cellcolor{ABcol}\ours{19.1} & \cellcolor{ABcol}--
& \cellcolor{ABcol}\ours{48.2} & \cellcolor{ABcol}-- \\
\midrule
Qwen3-8B
& 29.0 & 78.2\%
& 23.1 & 68.5\%
& 19.2 & 78.2\%
& 14.9 & 72.5\%
& 44.3 & 79.5\% \\
+ R-Zero
& 29.4 & 80.3\%
& 22.2 & 67.5\%
& 18.8 & 83.9\%
& 14.9 & 70.4\%
& 44.8 & 80.0\% \\
+ Absolute Zero
& 32.5 & 74.2\%
& 23.2 & 68.3\%
& 19.1 & 78.8\%
& 15.8 & 68.2\%
& 46.2 & 70.4\% \\
\rule{0pt}{8pt}
\cellcolor{ABcol}\textbf{\ours{+ SpyRL}}
& \cellcolor{ABcol}\ours{34.1} & \cellcolor{ABcol}--
& \cellcolor{ABcol}\ours{25.8} & \cellcolor{ABcol}--
& \cellcolor{ABcol}\ours{23.2} & \cellcolor{ABcol}--
& \cellcolor{ABcol}\ours{19.1} & \cellcolor{ABcol}--
& \cellcolor{ABcol}\ours{48.5} & \cellcolor{ABcol}-- \\
\bottomrule[1.5pt]
\end{tabular}
}
\end{table*}

\begin{figure}[t]
\vspace{-9pt}
	\centering
	\begin{minipage}{0.98\linewidth}
		\centerline{\includegraphics[width=\textwidth]{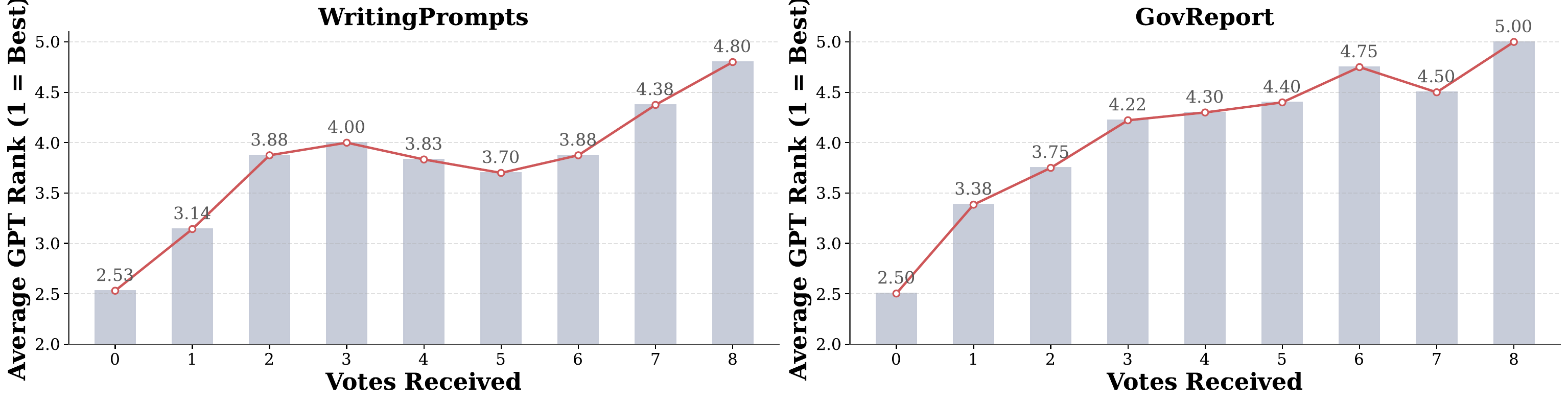}}
	\end{minipage}
    \vspace{-10pt}
	\caption{\textbf{Correlation between suspicion votes received in the detection stage and output quality as ranked by GPT-4o} ($1=\text{best}$, $5=\text{worst}$), measured over 100 games on WritingPrompts (left) and GovReport (right). Players receiving more votes consistently produce lower-quality outputs, confirming that the vote-based reward in SpyRL is well-aligned with actual task performance without requiring an external verifier.}
	\label{fig:vote_trend}
    \vspace{5pt}
\end{figure}

\vspace{-5pt}
\subsection{Main Results}
\vspace{-5pt}
SpyRL is universally compatible with both verifiable and unverifiable tasks. We present the performance of models trained on the three distinct categories of tasks in Tables~\ref{tab:summarization_results}--\ref{tab:main_results}.

\textbf{From Non-verifiable Reward to Verifiable Reward.}
To demonstrate that SpyRL can reliably transform non-verifiable rewards into verifiable training signals, we conduct an additional validation experiment. As shown in Figure~\ref{fig:vote_trend}, we run 100 games for both summarization and creative writing, and record the number of votes each player receives in the detection stage, \emph{i.e.}, the number of times the player is suspected of being the spy.
At the same time, for each game, we ask GPT-4o to rank the outputs of the five players in the performing stage by quality from best to worst ($1=\text{best}$, $5=\text{worst}$), and collect the rank of each player. We then compute the average GPT-4o rank for players receiving different numbers of votes. As illustrated in Figure~\ref{fig:vote_trend}, the number of votes a player receives is positively correlated with its rank number. In other words, lower-quality outputs tend to attract more suspicion votes and thus receive smaller rewards during training; conversely, higher-quality outputs tend to receive fewer votes and obtain larger rewards.
These results show that, in SpyRL, the reward assigned in the performing stage is directly aligned with task performance, enabling continuous performance improvement throughout training. Importantly, this improvement does not rely on any external verifier. Instead, SpyRL achieves it by cleverly converting a non-verifiable quality objective into a verifiable identity-discrimination problem.

\begin{table*}[t]
\centering
\setlength{\tabcolsep}{4.5pt}
\caption{\textbf{Results on creative writing benchmarks}. We report GPT-4o pairwise win rates (\%)
of SpyRL against the untrained backbone and against each self-evolution baseline.
A/B win rates of the baselines against their own backbone are
reported in Appendix~\ref{app:vs_base}.}
\label{tab:additional_abtest_writing}
\vspace{-5pt}
\resizebox{\textwidth}{!}{
\begin{tabular}{lccccc ccccc}
\toprule[1.5pt]
& \multicolumn{5}{c}{\textbf{WritingPrompt}} & \multicolumn{5}{c}{\textbf{WritingBench}} \\
\cmidrule(lr){2-6} \cmidrule(lr){7-11}
\textbf{SpyRL vs.}
& \shortstack{Novel}
& \shortstack{Emotion}
& \shortstack{Coher.}
& \shortstack{Consist.}
& \shortstack{Overall}
& \shortstack{Novel}
& \shortstack{Emotion}
& \shortstack{Coher.}
& \shortstack{Consist.}
& \shortstack{Overall} \\
\midrule
\midrule
\multicolumn{11}{c}{\textit{Qwen3-4B}} \\
\cellcolor{ABcol}\ours{Qwen3-4B}
& \cellcolor{ABcol}84.3\% & \cellcolor{ABcol}76.8\% & \cellcolor{ABcol}72.3\% & \cellcolor{ABcol}70.1\% & \cellcolor{ABcol}\ours{81.3\%}
& \cellcolor{ABcol}76.2\% & \cellcolor{ABcol}75.7\% & \cellcolor{ABcol}68.5\% & \cellcolor{ABcol}68.0\% & \cellcolor{ABcol}\ours{75.1\%} \\
\cellcolor{ABcol}\ours{R-Zero}
& \cellcolor{ABcol}86.2\% & \cellcolor{ABcol}80.3\% & \cellcolor{ABcol}74.2\% & \cellcolor{ABcol}74.8\% & \cellcolor{ABcol}\ours{78.9\%}
& \cellcolor{ABcol}76.4\% & \cellcolor{ABcol}79.3\% & \cellcolor{ABcol}72.8\% & \cellcolor{ABcol}71.4\% & \cellcolor{ABcol}\ours{75.0\%} \\
\cellcolor{ABcol}\ours{Absolute Zero}
& \cellcolor{ABcol}80.8\% & \cellcolor{ABcol}78.1\% & \cellcolor{ABcol}72.9\% & \cellcolor{ABcol}70.5\% & \cellcolor{ABcol}\ours{75.6\%}
& \cellcolor{ABcol}75.0\% & \cellcolor{ABcol}75.5\% & \cellcolor{ABcol}68.0\% & \cellcolor{ABcol}65.7\% & \cellcolor{ABcol}\ours{71.1\%} \\
\midrule
\multicolumn{11}{c}{\textit{Qwen3-8B}} \\
\cellcolor{ABcol}\ours{Qwen3-8B}
& \cellcolor{ABcol}77.3\% & \cellcolor{ABcol}76.2\% & \cellcolor{ABcol}74.2\% & \cellcolor{ABcol}75.0\% & \cellcolor{ABcol}\ours{76.5\%}
& \cellcolor{ABcol}78.1\% & \cellcolor{ABcol}77.4\% & \cellcolor{ABcol}71.0\% & \cellcolor{ABcol}71.2\% & \cellcolor{ABcol}\ours{78.1\%} \\
\cellcolor{ABcol}\ours{R-Zero}
& \cellcolor{ABcol}78.2\% & \cellcolor{ABcol}76.8\% & \cellcolor{ABcol}78.0\% & \cellcolor{ABcol}76.9\% & \cellcolor{ABcol}\ours{77.5\%}
& \cellcolor{ABcol}80.5\% & \cellcolor{ABcol}78.9\% & \cellcolor{ABcol}73.6\% & \cellcolor{ABcol}74.8\% & \cellcolor{ABcol}\ours{77.0\%} \\
\cellcolor{ABcol}\ours{Absolute Zero}
& \cellcolor{ABcol}70.4\% & \cellcolor{ABcol}78.1\% & \cellcolor{ABcol}70.4\% & \cellcolor{ABcol}70.5\% & \cellcolor{ABcol}\ours{72.4\%}
& \cellcolor{ABcol}76.4\% & \cellcolor{ABcol}75.3\% & \cellcolor{ABcol}68.9\% & \cellcolor{ABcol}68.2\% & \cellcolor{ABcol}\ours{72.2\%} \\
\bottomrule[1.5pt]
\end{tabular}
}
\end{table*}

\begin{table*}[t]
\vspace{-5pt}
\centering
\caption{\textbf{Results on mathematical and general reasoning benchmarks}. We report accuracy (\%)
on five mathematical benchmarks and two broader
reasoning benchmarks.}
\vspace{-5pt}
\label{tab:main_results}
\resizebox{\textwidth}{!}{
\begin{tabular}{lccccccc}
\toprule[1.5pt]
\textbf{Method} & \textbf{GSM8K} & \textbf{Math500} & \textbf{AIME 24} & \textbf{AIME 25} & \textbf{Minerva} & \textbf{MMLU-Pro} & \textbf{GPQA-D} \\
\midrule
Qwen3-4B
& 84.5 & 68.2 & 10.3 & 6.7 & 42.3 & 51.6 & 26.3 \\
+ R-Zero
& 88.7 & 72.8 & 10.3 & 6.7 & 47.1 & 52.8 & 27.8 \\
+ Absolute Zero
& 89.3 & 76.2 & 12.2 & 13.4 & 41.9 & 52.6 & 35.3 \\
\rule{0pt}{8pt}
\cellcolor{OursRow}\ours{+ SpyRL}
& \cellcolor{OursRow}\ours{93.4}
& \cellcolor{OursRow}\ours{79.5}
& \cellcolor{OursRow}\ours{13.3}
& \cellcolor{OursRow}\ours{20.0}
& \cellcolor{OursRow}\ours{47.8}
& \cellcolor{OursRow}\ours{57.4}
& \cellcolor{OursRow}\ours{41.3} \\
\midrule
Qwen3-8B
& 91.8 & 74.2 & 15.3 & 12.1 & 49.3 & 58.1 & 33.3 \\
+ R-Zero
& 92.1 & 78.4 & 15.3 & 14.2 & 52.5 & 61.7 & 34.3 \\
+ Absolute Zero
& 92.0 & 76.6 & 18.4 & 18.2 & 52.9 & 62.5 & 36.8 \\
\rule{0pt}{8pt}
\cellcolor{OursRow}\ours{+ SpyRL}
& \cellcolor{OursRow}\ours{93.5}
& \cellcolor{OursRow}\ours{81.2}
& \cellcolor{OursRow}\ours{20.0}
& \cellcolor{OursRow}\ours{23.3}
& \cellcolor{OursRow}\ours{56.3}
& \cellcolor{OursRow}\ours{63.1}
& \cellcolor{OursRow}\ours{39.8} \\
\bottomrule[1.5pt]
\end{tabular}
\vspace{20pt}
}
\end{table*}

\begin{table*}[t]
\centering
\setlength{\tabcolsep}{4.5pt}
\caption{\textbf{Human evaluation results on creative writing.} We report the pairwise win rates of SpyRL against each baseline across five evaluation dimensions. Higher is better.}
\vspace{-10pt}
\resizebox{\textwidth}{!}{
\begin{tabular}{lccccc ccccc}
\toprule[1.5pt]
& \multicolumn{5}{c}{\textbf{WritingPrompt}} & \multicolumn{5}{c}{\textbf{WritingBench}} \\
\cmidrule(lr){2-6} \cmidrule(lr){7-11}
\textbf{SpyRL vs}
& \shortstack{Novel}
& \shortstack{Emotion}
& \shortstack{Coher.}
& \shortstack{Consist.}
& \shortstack{Overall}
& \shortstack{Novel}
& \shortstack{Emotion}
& \shortstack{Coher.}
& \shortstack{Consist.}
& \shortstack{Overall} \\
\midrule
\midrule
\cellcolor{OursRow}\textbf{Qwen3-4B}
& \cellcolor{OursRow}80.0\% & \cellcolor{OursRow}82.5\% & \cellcolor{OursRow}70.5\% & \cellcolor{OursRow}64.5\% & \cellcolor{OursRow}80.0\%
& \cellcolor{OursRow}84.5\% & \cellcolor{OursRow}85.0\% & \cellcolor{OursRow}78.5\% & \cellcolor{OursRow}73.5\% & \cellcolor{OursRow}85.0\% \\

\cellcolor{OursRow}\textbf{RZero}
& \cellcolor{OursRow}77.5\% & \cellcolor{OursRow}84.0\% & \cellcolor{OursRow}68.0\% & \cellcolor{OursRow}66.0\% & \cellcolor{OursRow}78.5\%
&\cellcolor{OursRow} 81.5\% & \cellcolor{OursRow}80.0\% & \cellcolor{OursRow}72.5\% & \cellcolor{OursRow}70.0\% & \cellcolor{OursRow}80.5\% \\

\cellcolor{OursRow}\textbf{AbsoluteZero}
& \cellcolor{OursRow}73.0\% & \cellcolor{OursRow}80.5\% & \cellcolor{OursRow}64.5\% & \cellcolor{OursRow}67.5\% & \cellcolor{OursRow}74.0\%
& \cellcolor{OursRow}75.5\% & \cellcolor{OursRow}70.0\% & \cellcolor{OursRow}68.5\% & \cellcolor{OursRow}65.0\% & \cellcolor{OursRow}72.0\% \\
\bottomrule[1.5pt]
\end{tabular}
}
\label{tab:human_evaluation}
\end{table*}

\textbf{Performance on Non-verifiable Tasks.}
The results on non-verifiable tasks are presented in Tables~\ref{tab:summarization_results} and~\ref{tab:additional_abtest_writing}. Overall, SpyRL achieves the best performance on both summarization and creative writing, two representative open-ended generation tasks.
On summarization, SpyRL attains the highest ROUGE-L on every benchmark and for both backbones, improving over Absolute Zero on GovReport from 33.2 to 36.7 with Qwen3-4B and from 32.5 to 34.1 with Qwen3-8B.
The ABTest result show that SpyRL wins the majority of comparisons in all thirty cells.
The same pattern holds on creative writing (Table~\ref{tab:additional_abtest_writing}), where SpyRL is preferred in every fine-grained dimension against the backbone and both baselines. The margins are largest in novelty and emotion, indicating that the gains are not confined to surface fluency or structural regularity but also extend to more subjective aspects of open-ended generation.
The two baselines, in contrast, gain far less from their own self-evolution procedure. This is because these methods rely heavily on task verifiability: they depend on verifiable solver feedback to dynamically adjust difficulty. SpyRL, however, reformulates the quality objective as an identity-recognition problem, thereby encouraging the model to produce outputs that are consistently stronger and more convincing than those of other players. This design allows SpyRL to deliver significant gains on open-ended tasks.

\textbf{Performance on verifiable Tasks.}
The results on verifiable tasks are shown in Table~\ref{tab:main_results}. Overall, SpyRL again achieves the best performance on mathematical and verifiable reasoning tasks.
For mathematics, SpyRL achieves the best performance across all five benchmarks for both Qwen3-4B and Qwen3-8B. The gains are particularly pronounced on more challenging benchmarks such as AIME24 and AIME25. Meanwhile, SpyRL also achieves the best results on broader reasoning benchmarks such as MMLU-Pro and GPQA-D, suggesting that its advantage is not limited to pure mathematical problem solving, but generalizes to a wider range of verifiable reasoning scenarios.
Compared with its performance on non-verifiable tasks, R-Zero and Absolute Zero indeed show more noticeable gains on verifiable tasks.
Nevertheless, SpyRL still consistently outperforms these methods even in this setting. This suggests that the advantage of SpyRL does not merely stem from task verifiability itself. Rather, through reward transformation and within-group competition, SpyRL provides a finer-grained and more stable learning signal, encouraging the model to generate reasoning processes that are more rigorous, complete, and persuasive. In addition, compared with the conventional single proposer-solver paradigm~\citep{zhao2025absolute, huang2025r}, the group-based design of SpyRL mitigates the bias introduced by single-sample optimization, which further contributes to its strong gains on verifiable tasks.
We also report performance improvements on harder tasks in Appendix~\ref{app:harder_reasoning} (Table~\ref{tab:reasoning_benchmarks}).

\begin{table*}[t]
\vspace{-5pt}
\centering
\setlength{\tabcolsep}{4.5pt}
\caption{\textbf{A/B Test Comparison with rubric-as-reward baselines.} We report the pairwise win rates of SpyRL against Qwen3.5-27B-RaR and GPT-4o-RaR. SpyRL uses no external verifier, whereas Qwen3.5-27B-RaR and GPT-4o-RaR incur approximately \$200 and \$900 in additional verifier costs, respectively, in our experiments.}
\vspace{-10pt}
\resizebox{\textwidth}{!}{
\begin{tabular}{lccccc ccccc}
\toprule[1.5pt]
& \multicolumn{5}{c}{\textbf{WritingPrompt}} & \multicolumn{5}{c}{\textbf{WritingBench}} \\
\cmidrule(lr){2-6} \cmidrule(lr){7-11}
\textbf{SpyRL vs}
& \shortstack{Novel}
& \shortstack{Emotion}
& \shortstack{Coher.}
& \shortstack{Consist.}
& \shortstack{Overall}
& \shortstack{Novel}
& \shortstack{Emotion}
& \shortstack{Coher.}
& \shortstack{Consist.}
& \shortstack{Overall} \\
\midrule
\midrule
\cellcolor{OursRow}\textbf{Qwen3.5-RaR}
& \cellcolor{OursRow}\textbf{64.6\%} & \cellcolor{OursRow}\textbf{60.1\%} & \cellcolor{OursRow}\textbf{56.2\%} & \cellcolor{OursRow}\textbf{56.2\%} & \cellcolor{OursRow}\textbf{59.3\%}
& \cellcolor{OursRow}\textbf{58.4\%} & \cellcolor{OursRow}\textbf{59.2\%} & \cellcolor{OursRow}\textbf{54.4\%} & \cellcolor{OursRow}\textbf{52.8\%} & \cellcolor{OursRow}\textbf{56.2\%} \\
\midrule
\cellcolor{OursRow}\textbf{GPT-4o-RaR}
& \cellcolor{OursRow}\textbf{50.9\%} & \cellcolor{OursRow}\textbf{54.2\%} & \cellcolor{OursRow}45.8\% & \cellcolor{OursRow}44.5\% & \cellcolor{OursRow}48.9\%
& \cellcolor{OursRow}\textbf{51.8\%} & \cellcolor{OursRow}\textbf{52.2\%} & \cellcolor{OursRow}45.0\% & \cellcolor{OursRow}43.7\% & \cellcolor{OursRow}48.2\% \\

\bottomrule[1.5pt]
\end{tabular}
\label{tab:rar_results}
}
\vspace{10pt}
\end{table*}

\begin{table*}[t]

\centering
\setlength{\tabcolsep}{6pt}
\caption{\textbf{Results on scientific and domain-specific summarization benchmarks}. SpyRL is
trained on the PubMed corpus with the same game construction and hyperparameters as the main
summarization run, then evaluated on arXiv, PubMed, and BillSum. We report ROUGE-L (R-L) and
GPT-4o A/B test win rates (\%) against the untrained base model. The shaded row is SpyRL; the
\textbf{Average} columns are means over the three benchmarks.}
\label{tab:domain_summarization}
\vspace{-5pt}
\resizebox{1.0\textwidth}{!}{
\begin{tabular}{lcccccccc}
\toprule[1.5pt]
\multirow{2}{*}{\textbf{Method}}
& \multicolumn{2}{c}{\textbf{arXiv}}
& \multicolumn{2}{c}{\textbf{PubMed}}
& \multicolumn{2}{c}{\textbf{BillSum}}
& \multicolumn{2}{c}{\textbf{Average}} \\
\cmidrule(lr){2-3}
\cmidrule(lr){4-5}
\cmidrule(lr){6-7}
\cmidrule(lr){8-9}
& \textbf{ROUGE-L} & \textbf{ABTest}
& \textbf{ROUGE-L} & \textbf{ABTest}
& \textbf{ROUGE-L} & \textbf{ABTest}
& \textbf{ROUGE-L} & \textbf{ABTest} \\
\midrule
Qwen3-4B
& 28.1 & 52.8\%
& 30.3 & 51.2\%
& 41.3 & 54.8\%
& 33.2 & 52.9\% \\
\rule{0pt}{8pt}
\cellcolor{OursRow}\ours{+ SpyRL}
& \cellcolor{OursRow}\ours{32.5} & \cellcolor{OursRow}\ours{72.1\%}
& \cellcolor{OursRow}\ours{35.1} & \cellcolor{OursRow}\ours{68.9\%}
& \cellcolor{OursRow}\ours{46.8} & \cellcolor{OursRow}\ours{69.5\%}
& \cellcolor{OursRow}\ours{38.1} & \cellcolor{OursRow}\ours{70.2\%} \\
\bottomrule[1.5pt]
\end{tabular}}
\end{table*}

\textbf{Human Evaluation.}
We further conduct a blinded human evaluation on creative writing to assess whether the gains of SpyRL align with human preferences.
Ten Ph.D.\ students evaluate 400 randomly sampled prompts, with 200 from WritingPrompts and 200 from WritingBench; each evaluator assesses 40 instances. For each prompt, evaluators rank four anonymized responses from Qwen3-4B, R-Zero, Absolute Zero, and SpyRL across five dimensions. We report the pairwise win rate of SpyRL whenever its response is ranked above the corresponding baseline. As shown in Table~\ref{tab:human_evaluation}, SpyRL achieves overall win rates of 80.0\%, 78.5\%, and 74.0\% against Qwen3-4B, R-Zero, and Absolute Zero on WritingPrompts, respectively, with similarly strong results on WritingBench.
SpyRL also consistently leads across all fine-grained dimensions, particularly in novelty and emotion. These results confirm that the improvements of SpyRL are recognized by human evaluators.
We also report the agreement between human and LLM evaluations in Appendix~\ref{app:gpt_human_agreement} (Table~\ref{tab:gpt_human_agreement}).

\textbf{Comparison with Rubric-as-Reward Baselines.}
We further compare SpyRL with rubric-as-reward (RaR) methods that rely on external evaluators. Using the same GRPO framework, we train for 50 iterations with a batch size of 1024, employing Qwen3.5-27B and GPT-4o as rubric executors to construct Qwen3.5-27B-RaR and GPT-4o-RaR. As shown in Table~\ref{tab:rar_results}, SpyRL outperforms Qwen3.5-27B-RaR across all dimensions on both WritingPrompts and WritingBench, with overall win rates of 59.3\% and 56.2\%. Against the stronger GPT-4o-RaR, SpyRL remains competitive, while performing better in novelty and emotion. Moreover, SpyRL requires no external verifier, whereas Qwen3.5-27B-RaR and GPT-4o-RaR incur approximately \$200 and \$900 in additional verifier costs, respectively. These results highlight the cost-performance trade-off of SpyRL and the benefit of jointly improving performing and detection capabilities through self-play, without being constrained by fixed verifier.

\begin{table*}[t]
\vspace{-10pt}
\centering
\setlength{\tabcolsep}{6pt}
\caption{\textbf{Cross-task transfer results}. Each row takes the SpyRL model trained on one task
and evaluates it, without any further fine-tuning, on a task it was not trained on. We report the
GPT-4o A/B win rate (\%) against the corresponding untrained Qwen3-4B, so a value above $50\%$
indicates positive transfer ($\uparrow$) and a value below $50\%$ negative transfer ($\downarrow$).
Shaded rows are the positive-transfer cases.}
\label{tab:cross_task_transfer}
\vspace{-5pt}
\resizebox{\textwidth}{!}{
\begin{tabular}{lcccccc}
\toprule[1.5pt]
\multicolumn{7}{c}{Evaluated on Summarization benchmarks} \\
\midrule
\textbf{Trained on}
& \textbf{GovReport} & \textbf{Multi-News} & \textbf{QMSum}
& \textbf{VCSum} & \textbf{SAMSum} & \textbf{Trend} \\
\midrule

\cellcolor{OursRow}\ours{Creative Writing}
& \cellcolor{OursRow}\ours{56.1\%} & \cellcolor{OursRow}\ours{55.4\%} & \cellcolor{OursRow}\ours{52.7\%}
& \cellcolor{OursRow}\ours{51.8\%} & \cellcolor{OursRow}\ours{52.9\%} & \cellcolor{OursRow}\ours{$\uparrow$} \\
Mathematical Reasoning
& 45.6\% & 43.8\% & 44.1\% & 41.7\% & 42.5\% & $\downarrow$ \\
\midrule[1.2pt]
\multicolumn{7}{c}{Evaluated on Creative Writing benchmarks} \\
\midrule
\textbf{Trained on}
& \textbf{Novelty} & \textbf{Emotion} & \textbf{Coherence}
& \textbf{Consistency} & \textbf{Overall} & \textbf{Trend} \\
\midrule
\cellcolor{OursRow}\ours{Summarization}
& \cellcolor{OursRow}\ours{55.8\%} & \cellcolor{OursRow}\ours{53.2\%} & \cellcolor{OursRow}\ours{64.2\%}
& \cellcolor{OursRow}\ours{63.1\%} & \cellcolor{OursRow}\ours{59.1\%} & \cellcolor{OursRow}\ours{$\uparrow$} \\
Mathematical Reasoning
& 38.5\% & 40.7\% & 42.5\% & 42.1\% & 40.9\% & $\downarrow$ \\
\bottomrule[1.5pt]
\end{tabular}
}
\vspace{3pt}
\end{table*}

\begin{table}[t]
\centering
\caption{\textbf{Ablation study on Math500 accuracy (\%) across training epochs.} ``Only Performing'' freezes the detection stage; ``Only Detection'' freezes the performing stage; ``Without spy'' removes information asymmetry. SpyRL with full two-stage coupled optimization.}
\vspace{-10pt}
\label{tab:ablation_results}
\resizebox{\textwidth}{!}{
\begin{tabular}{lcccccc}
\toprule[1.5pt]
\textbf{Setting} & \textbf{Epoch = 0} & \textbf{Epoch = 20} & \textbf{Epoch = 40} & \textbf{Epoch = 60} & \textbf{Epoch = 80} & \textbf{Epoch = 100} \\
\midrule
Only Performing     & 68.2 & 70.1 & 71.8 & 72.2 & 71.9 & 72.3 \\
Only Detection & 68.2 & 69.0 & 69.4 & 69.0 & 69.2 & 69.2 \\
Without spy   & 68.2 & 69.6 & 71.1 & 69.8 & 70.5 & 71.6 \\
\cellcolor{OursRow}\textbf{SpyRL} & \cellcolor{OursRow}\textbf{68.2} & \cellcolor{OursRow}\textbf{73.3} & \cellcolor{OursRow}\textbf{78.2} & \cellcolor{OursRow}\textbf{78.8} & \cellcolor{OursRow}\textbf{79.3} & \cellcolor{OursRow}\textbf{79.5} \\
\bottomrule[1.5pt]
\end{tabular}}
\end{table}

\begin{figure}[t]
\vspace{-10pt}
	\centering
	\begin{minipage}{0.98\linewidth}
		\centerline{\includegraphics[width=\textwidth]{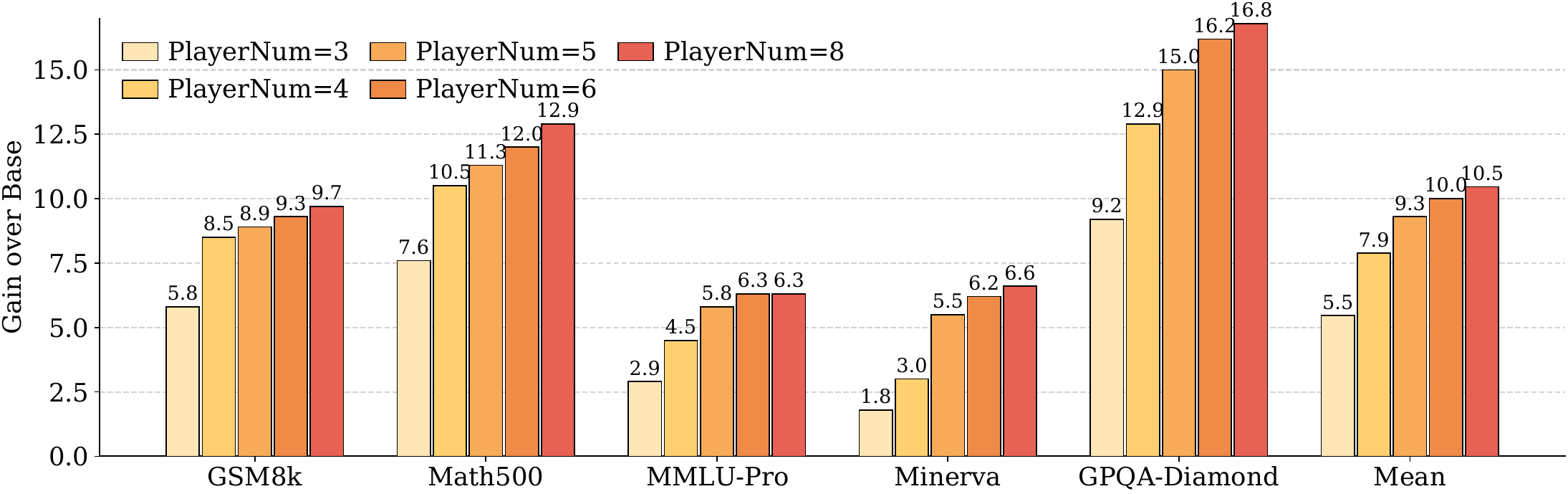}}
	\end{minipage}
    \vspace{-5pt}
	\caption{\textbf{Effect of group size ($n$) on SpyRL performance gain} over the base model across five reasoning benchmarks. Increasing the number of players from 3 to 5 yields the largest marginal improvement (mean gain: $5.5 \to 9.3$), while further scaling to 6 and 8 players shows diminishing returns, suggesting $n{=}5$ provides a sufficient game complexity.}
	\label{fig:playernum}
    \vspace{3pt}
\end{figure}

\subsection{Generalization Beyond the Main Benchmarks}

\textbf{Domain-Specific Summarization.}
The summarization experiments above all draw their training corpus from government reports, which leaves open whether SpyRL depends on that particular document distribution. We therefore repeat the summarization setting with scientific articles, training Qwen3-4B on the PubMed summarization corpus while keeping the game construction, prompts, and optimization settings unchanged: civilians receive the complete article, the spy receives the article with a continuous $20\%$ span masked, and the detector identifies whose summary was most likely written from incomplete source material. Evaluated on arXiv, PubMed, and BillSum (Table~\ref{tab:domain_summarization}), SpyRL raises ROUGE-L from 28.1 to 32.5, from 30.3 to 35.1, and from 41.3 to 46.8, an average gain of 4.9 points, with A/B win rates of 72.1\%, 68.9\%, and 69.5\%. The benefit of information-asymmetric self-play therefore carries over to scientific and other domain-specific sources.

\textbf{Cross-Task Transfer.}
We also ask whether the capabilities acquired in one domain carry over to another. We take each final checkpoint, evaluate it on a task it was not trained on, and compare it against the corresponding untrained Qwen3-4B under the same GPT-4o A/B protocol. As shown in Table~\ref{tab:cross_task_transfer}, summarization and creative writing transfer positively in both directions. This is expected, since both tasks draw on shared capabilities such as content organization, completeness, discourse coherence, and long-range consistency. The mathematical-reasoning model, by contrast, falls below parity on both writing tasks; it primarily strengthens symbolic manipulation and multi-step reasoning, which overlap far less with the stylistic and discourse-level demands of open-ended writing.

\begin{table*}[t]
\vspace{-20pt}
\centering
\setlength{\tabcolsep}{6pt}
\caption{\textbf{Ablation on Role-Advantage Estimation}. Accuracy (\%) of Qwen3-4B trained with and
without RAE, using identical data, reward function, and optimization settings; the w/o RAE variant
uses the raw performing rewards without subtracting the role-specific baseline.}
\label{tab:rae_reasoning_results}
\vspace{-5pt}
\resizebox{\textwidth}{!}{
\begin{tabular}{lcccccccc}
\toprule[1.5pt]
\textbf{Method} & \textbf{GSM8K} & \textbf{Math500} & \textbf{AIME 24} & \textbf{AIME 25}
& \textbf{Minerva} & \textbf{MMLU-Pro} & \textbf{GPQA-D} & \textbf{Avg.} \\
\midrule
Qwen3-4B
& 84.5 & 68.2 & 10.3 & 6.7 & 42.3 & 51.6 & 26.3 & 41.4 \\

+ SpyRL (w/o RAE)
& 76.1 & 61.9 & 10.3 & 6.7 & 36.2 & 46.5 & 25.1 & 37.5 \\

\cellcolor{OursRow}\ours{+ SpyRL (w/ RAE)}
& \cellcolor{OursRow}\ours{93.4} & \cellcolor{OursRow}\ours{79.5} & \cellcolor{OursRow}\ours{13.3}
& \cellcolor{OursRow}\ours{20.0} & \cellcolor{OursRow}\ours{47.8} & \cellcolor{OursRow}\ours{57.4}
& \cellcolor{OursRow}\ours{41.3} & \cellcolor{OursRow}\ours{50.4} \\
\bottomrule[1.5pt]
\end{tabular}}
\vspace{3pt}
\end{table*}

\begin{table*}[t]
\vspace{-10pt}
\centering
\setlength{\tabcolsep}{4pt}
\caption{\textbf{Sensitivity to the information-degradation operator $g(\cdot)$}. ROUGE-L and
GPT-4o A/B win rates (\%) against the untrained base model for SpyRL trained with a continuous
span-masking ratio of $20\%$ and $40\%$, with all other settings unchanged.}
\label{tab:summarization_ratio_results}
\vspace{-5pt}
\resizebox{\textwidth}{!}{
\begin{tabular}{lcccccccccc}
\toprule[1.5pt]
\multirow{2}{*}{\textbf{Model}}
& \multicolumn{2}{c}{\textbf{GovReport}}
& \multicolumn{2}{c}{\textbf{MultiNews}}
& \multicolumn{2}{c}{\textbf{QMSum}}
& \multicolumn{2}{c}{\textbf{VCSum}}
& \multicolumn{2}{c}{\textbf{SAMSum}} \\
\cmidrule(lr){2-3}
\cmidrule(lr){4-5}
\cmidrule(lr){6-7}
\cmidrule(lr){8-9}
\cmidrule(lr){10-11}
& \textbf{ROUGE-L} & \textbf{ABTest}
& \textbf{ROUGE-L} & \textbf{ABTest}
& \textbf{ROUGE-L} & \textbf{ABTest}
& \textbf{ROUGE-L} & \textbf{ABTest}
& \textbf{ROUGE-L} & \textbf{ABTest} \\
\midrule
Qwen3-4B
& 30.2 & 51.2\%
& 23.1 & 52.1\%
& 21.3 & 52.4\%
& 15.1 & 51.8\%
& 43.2 & 50.9\% \\

\cellcolor{OursRow}+ SpyRL (ratio $=20\%$)
& \cellcolor{OursRow}36.7 & \cellcolor{OursRow}74.6\%
& \cellcolor{OursRow}\ours{26.4} & \cellcolor{OursRow}\ours{80.2\%}
& \cellcolor{OursRow}\ours{25.3} & \cellcolor{OursRow}68.4\%
& \cellcolor{OursRow}19.1 & \cellcolor{OursRow}70.2\%
& \cellcolor{OursRow}48.2 & \cellcolor{OursRow}76.2\% \\
\cellcolor{OursRow}+ SpyRL (ratio $=40\%$)
& \cellcolor{OursRow}\ours{37.0} & \cellcolor{OursRow}\ours{75.8\%}
& \cellcolor{OursRow}25.8 & \cellcolor{OursRow}77.3\%
& \cellcolor{OursRow}24.8 & \cellcolor{OursRow}\ours{68.9\%}
& \cellcolor{OursRow}\ours{19.2} & \cellcolor{OursRow}\ours{72.5\%}
& \cellcolor{OursRow}\ours{49.1} & \cellcolor{OursRow}\ours{78.1\%} \\
\bottomrule[1.5pt]
\end{tabular}}
\vspace{3pt}
\end{table*}

\subsection{Ablation Study}\label{sec:ablation}

In this section, we conduct ablation studies to systematically disentangle the contributions of individual SpyRL components, alongside key hyperparameter analysis.

\textbf{Ablation on Modules.}
Table~\ref{tab:ablation_results} demonstrates that alternating optimization between the performing and detection stages is crucial.
Training exclusively on the performing stage initially exhibits noticeable improvements by directly providing task-correlated rewards.
However, as the model's intrinsic capabilities grow, maintaining a static detection module inevitably leads to reward distortion.
The frozen detector struggles to accurately discern quality nuances among increasingly sophisticated responses, causing the training process to rapidly plateau and oscillate.
Conversely, training solely on the detection stage yields negligible target-task improvements.
Furthermore, removing the spy mechanism yields a similar bottleneck phenomenon. Although the model achieves initial gains, its overall performance quickly stagnates. This underscores that without the auxiliary supervision and adversarial signals provisioned by the spy, the detection module cannot continuously co-evolve. Consequently, it fails to supply sufficiently accurate rewards to the performing stage.
We also report an ablation study of alternating training in Appendix~\ref{app:joint_vs_alternating}, Table~\ref{tab:joint_alternating}.

\textbf{Ablation on Group Size.} As shown in Figure~\ref{fig:playernum}, as the number of players ($n$) increases, the performance of SpyRL improves consistently across tasks. This suggests that enlarging the group size elevates the complexity and difficulty of the adversarial environment, thereby generating more rigorous and effective training signals that augment model performance. 
Notably, scaling the player count from 3 to 5 yields the most pronounced marginal gain, with the average performance surging from 5.5 to 9.3. As the player count further expands to 6 and 8, the performance improvements plateau, exhibiting a trend of diminishing marginal returns. This indicates that a group size of 5 already provides a sufficiently complex game environment to drive robust self-improvement.

\textbf{Ablation on Role-Advantage Estimation.}
Because the spy and the civilians face structurally different task difficulty, their raw performing rewards are not directly comparable, which is what RAE is designed to correct. To verify that this calibration matters, we train a variant that is identical in data, reward function, and optimization settings, except that the raw rewards are used directly without subtracting the role-specific baseline. As shown in Table~\ref{tab:rae_reasoning_results}, removing RAE lowers the seven-benchmark average from 50.4 to 37.5, and degrades GSM8K, Math500, Minerva, MMLU-Pro, and GPQA-Diamond relative to the backbone. Without role-aware calibration, the optimizer conflates the spy's information disadvantage with poor policy quality, so the resulting gradient actively harms the model rather than merely slowing it down.

\textbf{Sensitivity to the Degradation Operator.}
Since $g(\cdot)$ is the only component of SpyRL that must be specified per task, we ablate how much its design matters by comparing continuous span-masking ratios of $20\%$ and $40\%$ for summarization while holding everything else fixed. As shown in Table~\ref{tab:summarization_ratio_results}, the two settings are nearly indistinguishable; neither ratio is uniformly better across the five benchmarks. This insensitivity is expected, because continuous masking removes task-relevant content while preserving topic and surface form, and RAE re-centers rewards on each role's own baseline, so a player receives a positive signal whenever it beats its role-specific expectation regardless of how hard the operator makes the game overall. In practice, $g(\cdot)$ therefore requires little task-specific engineering, provided that the asymmetry it induces is meaningful but non-degenerate.

\vspace{-10pt}
\section{Conclusion}
We propose RLSVR, a training paradigm that extends RLVR to open-ended 
tasks by bringing the task-transformation principle of self-supervised 
learning into reinforcement learning: instead of approximating an 
unverifiable quality objective with external judges or reward models, 
RLSVR transforms the task into a proxy environment whose 
environment-assigned latent variables make rewards verifiable by 
construction. We instantiate this paradigm with SpyRL, an 
information-asymmetric self-play game in which a predetermined spy 
identity turns output-quality assessment into a verifiable 
identity-recognition problem. Across summarization, creative writing, 
and mathematical reasoning, SpyRL consistently outperforms existing 
self-improvement methods on open-ended tasks and yields further gains 
even on already-verifiable ones, with vote-based rewards shown to align 
closely with human and LLM quality judgments. Beyond the specific game, 
our results suggest a broader takeaway: verifiability need not be an 
intrinsic property of a task, but can be engineered through task 
transformation---opening a path toward scalable, verifier-free 
self-improvement on general open-ended capabilities.



\newpage
\bibliography{colm2026_conference}
\bibliographystyle{colm2026_conference}

\newpage
\appendix
\section{Related Work}

\paragraph{Multi-Agent Debate and Interaction.}
The idea of using debate as a mechanism for AI alignment was proposed by \citet{irving2018ai}. Since then, multi-agent debate has been shown to improve factuality and reasoning~\citep{du2024improving}, evaluation quality~\citep{chan2023chateval}, divergent thinking~\citep{liang2024encouraging}, and truthfulness~\citep{khan2024debating} in LLMs. Social deduction games have also been used to train LLMs with multi-agent RL~\citep{sarkar2025training}. However, these approaches primarily operate at inference time or target specific game performance; few convert multi-agent outcomes into training rewards for general open-domain tasks. By contrast, SpyRL uses an information-asymmetric adversarial game with rule-based outcomes as a verifiable RL training signal for arbitrary open-domain tasks.

\paragraph{Self-Play for LLMs.}
Self-play has driven breakthroughs from AlphaGo~\citep{silver2016mastering,silver2017mastering} to AlphaZero~\citep{silver2018general} and OpenAI Five~\citep{berner2019dota}, and asymmetric self-play creates automatic curricula~\citep{sukhbaatar2017asymmetric}.
In LLMs, self-play has been adapted for self-improvement~\citep{chen2024spin, yuan2024self} and reasoning. Proposer-solver frameworks such as Absolute Zero~\citep{zhao2025absolute}, R-Zero~\citep{huang2025r}, and Tool-R0~\citep{acikgoz2026toolr0} jointly evolve task generators and solvers, often achieving strong performance without external data~\citep{kuba2025languageselfplay}. Multi-agent self-play methods, including SPIRAL~\citep{liu2025spiral}, SPICE~\citep{liu2025spice}, SPAG~\citep{cheng2024self}, and SPELL~\citep{yang2025spell}, target reasoning through competitive games. Concurrently, Vision-Zero~\citep{wang2025vision} extends the same ``Who Is the spy'' game structure to vision-language models, enabling label-free self-improvement from arbitrary image inputs.
These methods primarily focus on verifiable domains (math/code) or use game-specific outcomes as indirect signals for general capabilities, risking degeneration under prolonged training~\citep{chae2025understandingselfplay, shafayat2025largereasoningmodelsselftrain}. SpyRL extends self-play to non-verifiable domains via an information-asymmetric game explicitly designed so that game success requires target-task proficiency, yielding verifiable training signals without answer-level verification.

\paragraph{RL Beyond Verifiable Domains.}
RLHF~\citep{ouyang2022training, bai2022training} and its variants~\citep{bai2022constitutional, lee2024rlaif, rafailov2023direct} replace deterministic verification with preference signals, while LLM-as-a-Judge~\citep{zheng2023judging}, rubric-based rewards~\citep{gunjal2025rubrics}, and self-rewarding approaches~\citep{yuan2024self} approximate verifiable feedback for open-ended tasks. Process reward models~\citep{lightman2023let, cobbe2021training} provide step-level feedback but require problems with deterministic answers. Concurrently, Writing-Zero~\citep{jia2025writingzero} bridges non-verifiable tasks via generative reward models. Extending RL beyond verifiable domains thus requires either expensive human preferences, model-generated judgments that introduce bias, or ground-truth answers that limit domain coverage. SpyRL takes a different approach: it transforms the quality objective into a verifiable identity-recognition problem where spy identity provides deterministic ground truth, yielding rule-based rewards without learned verifiers.
\section{Extended Related Work}\label{app:related}

\paragraph{Multi-Agent Debate and Interaction for LLMs.}
Multi-agent systems have a rich history in game AI, from TD-Gammon~\citep{tesauro1995temporal} and AlphaGo~\citep{silver2016mastering,silver2017mastering} to AlphaStar~\citep{vinyals2019grandmaster} and emergent tool use from multi-agent autocurricula~\citep{baker2019emergent}. In language, multi-agent interaction has been explored through emergent communication~\citep{graesser2019emergent} and strategic reasoning in Diplomacy~\citep{meta2022human}.
The idea of using debate as a mechanism for AI alignment was proposed by \citet{irving2018ai}. Since then, multi-agent debate has been shown to improve factuality and reasoning in LLMs~\citep{du2024improving}, enhance evaluation quality via deliberation~\citep{chan2023chateval}, encourage divergent thinking~\citep{liang2024encouraging}, and produce more truthful answers through debate~\citep{khan2024debating}. Social deduction games such as Among Us have also been used to train LLMs with multi-agent RL~\citep{sarkar2025training}, and multi-agent meta-reasoning has been explored in REMA~\citep{wan2025rema}.
However, these approaches primarily operate at inference time to improve output quality, or target performance within specific game environments; few convert multi-agent interaction outcomes into training rewards for general open-domain tasks.
By contrast, SpyRL formulates multi-agent interaction as an information-asymmetric adversarial game with rule-based outcomes, converting the interaction signal into a verifiable RL training reward applicable to arbitrary open-domain tasks.

\paragraph{Self-Play Training for LLMs.}
Self-play has driven sustained capability scaling from AlphaZero~\citep{silver2018general} to OpenAI Five~\citep{berner2019dota}, and asymmetric self-play has been shown to create powerful automatic curricula~\citep{sukhbaatar2017asymmetric}. In the LLM era, RL with verifiable rewards (RLVR) at scale, exemplified by DeepSeek-R1~\citep{deepseek2025r1}, OpenAI o1~\citep{openai2024o1}, and Kimi k1.5~\citep{kimiteam2025kimi}, has demonstrated that rule-based rewards can unlock chain-of-thought reasoning~\citep{wei2022chain}.
Self-play has been adapted for LLM self-improvement through SPIN~\citep{chen2024spin} and Self-Rewarding Language Models~\citep{yuan2024self}, while SeRL~\citep{fang2025serl} combines self-instruction with self-rewarding under limited data. A series of recent proposer-solver frameworks, including Absolute Zero~\citep{zhao2025absolute}, R-Zero~\citep{huang2025r}, Self-Questioning LMs~\citep{chen2025selfquestioning}, and Tool-R0~\citep{acikgoz2026toolr0}, jointly evolve task generators and solvers via self-play, often achieving strong performance without external data~\citep{singh2023beyond, kuba2025languageselfplay}. Multi-agent self-play has also been explored for reasoning: SPIRAL~\citep{liu2025spiral} and SPICE~\citep{liu2025spice} leverage competitive games and corpus environments respectively, SPAG~\citep{cheng2024self} uses adversarial taboo, SPELL~\citep{yang2025spell} targets long-context evolution, SPC~\citep{chen2025spcevolvingselfplaycritic} evolves critics via adversarial games, and Prover-Verifier Games~\citep{kirchner2024proververifiergamesimprovelegibility} improve output legibility through adversarial training. Concurrently, Vision-Zero~\citep{wang2025vision} extends the ``Who Is the spy'' game structure to vision-language models, demonstrating that the same self-play paradigm enables label-free VLM self-improvement from arbitrary image inputs.
However, these methods primarily target verifiable domains (math/code) or use game-specific outcomes as indirect signals for general capabilities, which limits applicability to open-ended tasks where objectives resist simple rule-based evaluation. Recent analyses further show that prolonged self-play risks degeneration and performance collapse~\citep{chae2025understandingselfplay, shafayat2025largereasoningmodelsselftrain}.
SpyRL extends self-play to non-verifiable domains by designing an information-asymmetric adversarial game in which game success directly requires target-task proficiency, yielding verifiable training signals without answer-level verification.

\paragraph{Reinforcement Learning Beyond Verifiable Domains.}
RLHF~\citep{ouyang2022training, bai2022training} and its variants, including Constitutional AI~\citep{bai2022constitutional}, RLAIF~\citep{lee2024rlaif}, and DPO~\citep{rafailov2023direct}, replace deterministic verification with learned or model-generated preference signals, but these approaches require expensive preference data or an external reward model whose quality bounds the learning. LLM-as-a-Judge~\citep{zheng2023judging} and rubric-based rewards~\citep{gunjal2025rubrics} attempt to approximate verifiable feedback for open-ended tasks, while self-rewarding approaches~\citep{yuan2024self} let the model serve as its own judge, coupling actor and evaluator capabilities.
Process reward models~\citep{lightman2023let, cobbe2021training} provide step-level feedback but require problems with deterministic ground-truth answers. Curriculum design~\citep{bengio2009curriculum}, unsupervised environment generation~\citep{dennis2020paired}, and scalable RL training systems~\citep{yu2025dapo} improve training efficiency but do not address the fundamental challenge of reward construction for unverifiable tasks.
Concurrently, Writing-Zero~\citep{jia2025writingzero} bridges non-verifiable creative writing tasks and verifiable rewards via principle-based generative reward models, and AlphaProof~\citep{AlphaProofNature2025} demonstrates that grounded RL with formal verification can produce complex mathematical reasoning strategies.
A common thread across these approaches is that extending RL beyond verifiable domains requires either expensive human preference data, model-generated quality judgments that introduce systematic bias, or ground-truth answers that limit domain coverage.
SpyRL takes a fundamentally different approach: rather than approximating task quality via human or model judgments, it transforms the quality objective into a verifiable identity-recognition problem through multi-agent competition. Because the spy identity is assigned by the environment, the correctness of detection votes is deterministically verifiable, yielding rule-based rewards that require neither external annotations nor learned verifiers.

\section{Implementation Details}\label{app:details}

\subsection{Role-Advantage Estimation}\label{app:rae}

In the performing stage, the spy player and the civilian players operate under structurally asymmetric information: the spy observes the degraded input $g(x)$ while civilians observe the full input $x$. This asymmetry induces systematically different expected reward distributions across the two roles, even when the underlying policy is identical. Directly using the raw performing rewards $r_u^P$ and $r_{c_j}^P$ for policy optimization would therefore conflate role-induced reward differences with genuine performance differences, leading to biased gradient estimates.

Following \citet{liu2025spiral}, we adopt Role-Advantage Estimation (RAE) to decouple role-specific reward baselines from the optimization signal. We maintain two exponential moving average (EMA) baselines, one for each role:
\begin{equation}
b_u \leftarrow \alpha \, b_u + (1-\alpha) \, r_u^P, \qquad
b_c \leftarrow \alpha \, b_c + (1-\alpha) \, \frac{1}{n_c}\sum_{j=1}^{n_c} r_{c_j}^P,
\end{equation}
where $\alpha \in [0,1)$ is the EMA decay rate and both baselines are initialized to zero. The baseline $b_u$ tracks the expected reward for the spy role, while $b_c$ tracks the expected reward for the civilian role. The role-calibrated advantages for the performing stage are then computed as:
\begin{equation}
A_u^P = r_u^P - b_u, \qquad A_{c_j}^P = r_{c_j}^P - b_c, \quad j = 1, \dots, n_c.
\end{equation}
These advantages $A_k^P$ (for $k \in \{u\} \cup \mathcal{C}$) replace the raw rewards in the performing-stage policy gradient (Equation~\ref{equ6} in the main text). By subtracting role-specific baselines, RAE ensures that the gradient signal reflects how well a player performed \emph{relative to the typical outcome for its assigned role}, rather than being confounded by the inherent difficulty difference between playing as the spy versus a civilian. This prevents the optimization from systematically favoring one role over the other and stabilizes training throughout the alternating optimization process.

\begin{table*}[t]
\centering
\vspace{-25pt}
\caption{\textbf{Parameter settings for state transitions.}
We report the thresholds used for transitions between the Detection and Performing states, together with the minimum dwell time.}
\label{tab:transition_parameters}
\vspace{-5pt}
\resizebox{0.7\textwidth}{!}{
\begin{tabular}{llc}
\toprule[1.5pt]
\textbf{Parameter} & \textbf{Description} & \textbf{Value} \\
\midrule
\midrule
\texttt{T\_acc}$\uparrow$
& accuracy threshold for Detection $\rightarrow$ Performing
& 0.9 \\
\midrule
\texttt{T\_err}$\uparrow$
& error threshold for Performing $\rightarrow$ Detection
& 0.4 \\
\midrule
\texttt{T\_na}$\uparrow$
& N/A threshold for Performing $\rightarrow$ Detection
& 0.5 \\
\midrule
\texttt{T\_na}$\downarrow$
& N/A threshold for Detection $\rightarrow$ Performing
& 0.1 \\
\midrule
\texttt{K\_min}
& minimum dwell time
& 5 \\
\bottomrule[1.5pt]
\end{tabular}}
\vspace{8pt}
\end{table*}

\subsection{Alternating Optimization Strategy}\label{app:alternating}
Pure self-play frameworks often suffer from local equilibria or knowledge saturation, where models merely exploit current game distributions rather than exploring novel reasoning paths. To sustain a challenging learning environment and ensure continuous co-evolution, SpyRL employs a dynamic, two-stage alternating training scheme between the \textit{Detection} and \textit{Performing} stages. 

Intuitively, if the detection stage easily identifies the spy player, the performing policy (specifically the spy's ability to blend in) is under-optimized and needs improvement. Conversely, if detection frequently fails or the model abstains, the detection policy is saturated and requires training. To formalize this, we monitor the detection performance over a mini-batch $\mathcal{B}_t$ at iteration $t$. We calculate the average identification accuracy $\mathrm{acc}_t$ and the uncertainty (``N/A'') rate $\mathrm{na}_t$. To prevent noisy gradient updates from triggering premature stage switches, we apply an exponential moving average (EMA) with a smoothing factor $\rho \in [0,1)$ to obtain stable estimates $\bar{\mathrm{acc}}_t$ and $\bar{\mathrm{na}}_t$.

Let $m_t \in \{0,1\}$ denote the active training phase, where $m_t=1$ activates the \textit{Performing} stage and $m_t=0$ activates the \textit{Detection} stage. We govern the phase transitions using a set of hysteresis thresholds ($\tau^{\uparrow}_{\mathrm{acc}}, \tau^{\uparrow}_{\mathrm{err}}, \tau^{\uparrow}_{\mathrm{na}}, \tau^{\downarrow}_{\mathrm{na}}$):

\begin{align}
\textbf{Detection} \rightarrow \textbf{Performing} \ (m_{t+1}=1):&\quad \text{if } m_t=0 \land \bar{\mathrm{acc}}_t \ge \tau^{\uparrow}_{\mathrm{acc}} \land \bar{\mathrm{na}}_t \le \tau^{\downarrow}_{\mathrm{na}}, \label{eq:switch-d2p} \\
\textbf{Performing} \rightarrow \textbf{Detection} \ (m_{t+1}=0):&\quad \text{if } m_t=1 \land \Big(1-\bar{\mathrm{acc}}_t \ge \tau^{\uparrow}_{\mathrm{err}} \lor \bar{\mathrm{na}}_t \ge \tau^{\uparrow}_{\mathrm{na}}\Big). \label{eq:switch-p2d}
\end{align}

If neither condition is met, the phase remains unchanged ($m_{t+1} = m_t$). To further avoid training chattering, we enforce a minimum dwell time of $K_{\min}$ updates per phase. The threshold values used in all experiments are listed in Table~\ref{tab:transition_parameters}. Under this gating mechanism, gradients are exclusively routed to the active module. 

This alternating paradigm provides two critical benefits: (1) It prevents the model from stagnating in a strategic equilibrium by dynamically switching stages based on real-time saturation signals, ensuring continuous adversarial improvement. (2) It introduces a stable supervision signal derived from verifiable game mechanics, preventing common multi-agent pitfalls such as role collapse or divergence.

\subsection{Training Hyperparameters}\label{app:hyperparams}

We optimize our model using the Group Relative Policy Optimization (GRPO) algorithm, implemented via the \texttt{verl} training framework. The base model for our actor and reference policies is \texttt{Qwen/Qwen3-4B-Instruct-2507}. Training was conducted on a single node equipped with 8 GPUs.

\begin{table}[H]
\centering
\small
\begin{tabular}{lc}
\toprule
\textbf{Hyperparameter} & \textbf{Value} \\
\midrule
\multicolumn{2}{l}{\textit{Algorithm \& Optimization}} \\
RL Algorithm & GRPO \\
Learning Rate & $1 \times 10^{-6}$ \\
Prompts per Batch & 128 \\
GRPO Rollouts per Prompt & 8 \\
Effective Batch Size & 1024 \\
PPO Mini-batch Size & 128 \\
Micro-batch Size (per GPU) & 2 \\
KL Penalty Coefficient ($\beta$) & 0.001 \\
KL Loss Type & Low-variance KL \\
Training Iterations & 100 \\
\midrule
\multicolumn{2}{l}{\textit{Length Constraints \& Environment}} \\
Max Prompt Length & 12,288 \\
Max Response Length & 4,096 \\
Max Model Length & 16,384 \\
Number of Players & 5 \\
Number of Rounds & 1 \\
\midrule
\multicolumn{2}{l}{\textit{System \& Memory}} \\
Hardware & 1 Node $\times$ 8 GPUs \\
Rollout Engine & vLLM (TP=8) \\
Gradient Checkpointing & True \\
Reference Model Offload & True \\
Actor Model Offload & False \\
\bottomrule
\end{tabular}
\caption{Key hyperparameters for the GRPO training phase.}
\label{tab:hyperparams}
\end{table}

\paragraph{Optimization and Algorithm Settings.} 
For the GRPO algorithm, we sample a group of $n=8$ responses per prompt during the rollout phase. With 128 prompts per batch and 8 rollouts each, this yields an effective batch size of 1024 samples. The actor model is trained with a learning rate of $1 \times 10^{-6}$ for 100 training iterations. The PPO mini-batch size is set to 128, with a micro-batch size of 2 per GPU. To prevent the policy from deviating excessively from the reference model, we apply a low-variance KL divergence penalty with a coefficient of $0.001$. The entropy coefficient is set to 0.

\paragraph{Generation and Rollout.} 
During the interactive rollout phase, we utilize the vLLM engine to accelerate generation, setting tensor model parallelism (TP) to 8 and restricting GPU memory utilization to 0.45 to leave sufficient memory for the training weights. We allow a maximum prompt length of 12,288 tokens and generate responses up to 4,096 tokens, bounded by a total maximum model length of 16,384 tokens. The game environment is configured for 5 players interacting over 1 round.

\paragraph{Memory Management and System Configurations.}
To manage GPU memory efficiently during the reinforcement learning process, we enable gradient checkpointing for the actor model. Furthermore, we employ Fully Sharded Data Parallel (FSDP). Specifically, the reference model's parameters are offloaded to the CPU (\texttt{param\_offload=True}) to save VRAM, while the actor model's parameters and optimizer states remain on the GPU to maximize training throughput.

Table \ref{tab:hyperparams} summarizes the key hyperparameter configurations used in our experiments.

The complete launch script and detailed configuration flags are provided below for reproducibility:

\begin{verbatim}
python3 -m verl.trainer.main_ppo \
    algorithm.adv_estimator=grpo \
    data.train_batch_size=128 \
    data.train_max_samples=100000000 \
    data.max_prompt_length=12288 \
    data.max_response_length=3762 \
    data.filter_overlong_prompts=True \
    +data.num_players=5 \
    +data.num_rounds=1 \
    +data.prompt_max_tokens=128 \
    actor_rollout_ref.rollout.agent.default_agent_loop=writingprompts_two_player \
    custom_reward_function.name=compute_score \
    +custom_reward_function.reward_kwargs.max_debug_prints=4 \
    actor_rollout_ref.model.path=Qwen/Qwen3-4B-Instruct-2507 \
    actor_rollout_ref.actor.optim.lr=1e-6 \
    actor_rollout_ref.actor.use_torch_compile=False \
    actor_rollout_ref.model.use_remove_padding=True \
    actor_rollout_ref.actor.ppo_mini_batch_size=128 \
    actor_rollout_ref.actor.ppo_micro_batch_size_per_gpu=2 \
    actor_rollout_ref.actor.use_kl_loss=True \
    actor_rollout_ref.actor.kl_loss_coef=0.001 \
    actor_rollout_ref.actor.kl_loss_type=low_var_kl \
    actor_rollout_ref.actor.entropy_coeff=0 \
    actor_rollout_ref.model.enable_gradient_checkpointing=True \
    actor_rollout_ref.actor.fsdp_config.param_offload=False \
    actor_rollout_ref.actor.fsdp_config.use_torch_compile=False \
    actor_rollout_ref.actor.fsdp_config.optimizer_offload=False \
    actor_rollout_ref.rollout.log_prob_micro_batch_size_per_gpu=2 \
    actor_rollout_ref.rollout.tensor_model_parallel_size=8 \
    actor_rollout_ref.rollout.name=vllm \
    actor_rollout_ref.rollout.gpu_memory_utilization=0.45 \
    actor_rollout_ref.rollout.enforce_eager=True \
    +actor_rollout_ref.rollout.engine_kwargs.vllm.compilation_config.cudagraph_mode=NONE \
    +actor_rollout_ref.rollout.engine_kwargs.vllm.compilation_config.use_inductor=False \
    actor_rollout_ref.rollout.agent.num_workers=1 \
    actor_rollout_ref.rollout.max_num_seqs=128 \
    actor_rollout_ref.rollout.max_num_batched_tokens=8192 \
    actor_rollout_ref.rollout.max_model_len=16384 \
    actor_rollout_ref.rollout.response_length=4096 \
    actor_rollout_ref.rollout.n=8 \
    actor_rollout_ref.ref.log_prob_micro_batch_size_per_gpu=2 \
    actor_rollout_ref.ref.use_torch_compile=False \
    actor_rollout_ref.ref.fsdp_config.use_torch_compile=False \
    actor_rollout_ref.ref.fsdp_config.param_offload=True \
    algorithm.use_kl_in_reward=False \
    trainer.critic_warmup=0 \
    trainer.val_before_train=False \
    +trainer.training_phase=interactive \
    +trainer.interactive_cycle_length=1 \
    trainer.logger='["console"]' \
    trainer.n_gpus_per_node=8 \
    trainer.nnodes=1 \
    trainer.save_freq=5 \
    trainer.test_freq=-1 \
    trainer.total_epochs=1 $@
\end{verbatim}

\subsection{Baselines \& Metrics}\label{app:baselines}
To rigorously evaluate the effectiveness of our approach, we benchmark against state-of-the-art self-play methodologies and employ a comprehensive evaluation protocol combining automated metrics with robust LLM-as-a-judge A/B testing. Prompt templates for the two game stages are provided in Appendix~\ref{app:prompt}.

\paragraph{Baselines.} 
We compare our model against two leading \textit{proposer-solver} self-play frameworks designed for large language models:
\begin{itemize}
    \item \textbf{R-Zero \citep{huang2025r}:} A state-of-the-art framework that leverages a self-contained reinforcement learning loop. In this paradigm, a \textit{proposer} model generates candidate solutions or trajectories, while a \textit{solver} (or reward model) evaluates them based on internal logic or rule-based constraints. This method eliminates the need for external human-annotated preference data by iteratively optimizing the proposer against the solver's feedback.
    \item \textbf{Absolute Zero \citep{zhao2025absolute}:} An advanced pure self-play framework inspired by zero-shot bootstrapping principles. It relies on iterative self-refinement where the model acts as both the generator and the verifier. Absolute Zero focuses on discovering optimal strategies entirely from the base model's inherent capabilities, providing a strong baseline for self-taught reasoning and strategic generation.
\end{itemize}
By comparing against these frameworks, we aim to demonstrate whether our specific multi-agent interactive training offers superior strategic adaptation and generation quality compared to standard single-agent or dual-agent proposer-solver loops.

\paragraph{Automatic Metrics.} 
Quantitative evaluation is first conducted using task-specific automated metrics. Depending on the exact nature of the game rounds, these metrics measure structural compliance, such as format adherence (e.g., successful extraction of the required \texttt{\textbackslash boxed\{\}} formatting), exact match accuracy for the detection stage (whether the spy is correctly identified), and basic linguistic metrics (e.g., word count constraints and repetition penalties). These rule-based metrics provide an objective baseline for the model's fundamental instruction-following capabilities.

\paragraph{GPT-4o A/B Testing and Position Bias Mitigation.} 
Because automated metrics fall short in assessing open-ended creativity, strategic depth, and narrative coherence (especially in the story-writing phase), we utilize GPT-4o as an impartial judge to conduct pairwise A/B testing. We compare the responses generated by our trained model directly against those from the base model and the baselines.

A well-known challenge in LLM-as-a-judge evaluation is \textit{position bias} (also known as order bias), where the evaluator disproportionately favors either the first or the second option presented in the prompt, regardless of actual quality. To rigorously mitigate this bias and ensure statistical significance, we implement a \textbf{swapped-order evaluation protocol}:
\begin{enumerate}
    \item For every comparison between Model A and Model B on a given prompt, we query GPT-4o twice.
    \item In the first query, Model A's output is presented as ``Candidate 1'' and Model B's as ``Candidate 2''.
    \item In the second query, the order is strictly reversed: Model B is presented as ``Candidate 1'' and Model A as ``Candidate 2''.
\end{enumerate}
We aggregate the pairwise results conservatively: Model A is only awarded a \textit{Win} if it is preferred in both permutations, or if it wins in one permutation and ties in the other. If Model A wins in the first query but loses in the swapped query (or vice versa), the result is recorded as a \textit{Tie}. This strict aggregation heavily penalizes position bias and ensures that any reported win rate reflects a genuine, robust preference for the generated content's quality.

\subsection{Prompt Design and Configurations}\label{app:prompt}

In our framework, the interactions are driven by two carefully engineered prompts corresponding to the two main phases of the game: the \textit{Performing Stage} and the \textit{Detection Stage}. These prompts are designed not merely to instruct the models, but to induce strategic reasoning, enforce high-quality text generation, and ensure robust automated parsing.

\subsubsection{Performing Stage Prompt Design}
The prompt for the Performing Stage acts as the system instruction for the agents generating the stories. Its key design advantages include:

\begin{itemize}
    \item \textbf{Asymmetric Role Awareness:} The prompt dynamically injects variables such as \texttt{\{role\_info\}} and \texttt{\{role\_instruction\}}, enabling the same base template to function for both civilian players and the spy. It explicitly emphasizes the information gap (the spy sees a blank prompt), establishing the core tension of the game.
    \item \textbf{Strategic Chain-of-Thought (CoT):} We require the agent to ``conduct your own thinking process'' before outputting the final answer. For the spy, this encourages reasoning about what the hidden prompt might be based on previous context; for civilian players, it encourages planning the narrative structure.
    \item \textbf{High-Quality Constraints:} To prevent standard, cliche LLM responses, the prompt provides explicit writing criteria (e.g., ``include subtext, a turning point,'' ``fresh imagery''). This forces the model to generate diverse and literary outputs rather than generic summaries.
    \item \textbf{Strict Formatting:} By mandating the exact prefix \texttt{"Answer:"} and restricting bullet points, we ensure the output is easily parsed by our automated evaluation pipeline.
\end{itemize}

\newtcblisting{promptbox}[1][]{
    colback=gray!5,           
    colframe=gray!40!black,   
    listing only,             
    breakable,                
    title=Performing Stage Prompt, 
    fonttitle=\bfseries,      
    sharp corners,            
    boxrule=0.8pt,            
    listing options={
        basicstyle=\ttfamily\small, 
        breaklines=true,            
        columns=fullflexible,
        keepspaces=true             
    },
    #1
}

\begin{promptbox}
game_context = f"""You are PLAYER {player_id} of the game "Who's the Odd One Out?". This is ROUND {round_num}. Your task is to write ONE short creative story based on the writing prompt.

{role_info}

Game Instructions:
- There are {num_players} players in total. Only one player (the spy) received NO writing prompt, while all other players received the SAME writing prompt.
- The player who is voted as the odd player will be severely punished, so you should convince others that you are not the odd player.
- The spy sees nothing (blank), while all other players see the same writing prompt.
{role_instruction}

What you should do:
1. Compare your writing prompt (or lack thereof) with all previous stories to understand what others are writing about.
2. If you are NOT the spy:
    1) Identify the core theme, mood, and implied constraints in the prompt.
    2) Write a short story that clearly matches the theme and is NOT generic.
    3) Make it deep: include subtext, a turning point, or a meaningful implication.
    4) Keep it coherent: clear setup -> development -> payoff, with smooth transitions.
    5) Be creative: fresh imagery, non-obvious choices, clever structure or constraints.
3. If you are the spy:
    1) Infer the likely theme/prompt by reading previous players' stories.
    2) Write a plausible story that fits what others are writing about.
    3) Your story can be slightly more general, but it should still feel thematic, coherent, and creative -- not obviously off-topic.

Fixed template you MUST follow:
- Story template: Write ONE continuous short story in plain paragraphs (no bullet points), with 300-500 words (strict). Include a clear turning point and a payoff.

Required output format:
First, conduct your own thinking process.
Then, start a new line with exactly "Answer:" and write exactly ONE short story following the Story template after it (may span multiple paragraphs).

{prompt_display}
"""
\end{promptbox}

\subsubsection{Detection Stage Prompt Design}
The Detection Stage prompt transforms the LLM into a critical evaluator. To minimize hallucination and subjective judging, we incorporate the following design features:

\begin{itemize}
    \item \textbf{Multi-dimensional Evaluation Rubric:} Instead of simply asking ``who is the spy,'' the prompt provides 5 specific dimensions of anomaly detection (e.g., \textit{Off-theme}, \textit{Shallow}, \textit{Consistency with others}). This grounds the model's reasoning in concrete evidence.
    \item \textbf{Uncertainty Calibration:} By explicitly permitting the output \texttt{\textbackslash boxed\{N/A\}}, we allow the model to express uncertainty. This prevents forced guesses when evidence is insufficient and mimics human hesitation, allowing the model to continue reasoning in subsequent iterations.
    \item \textbf{Efficiency and Length Control:} The prompt imposes a strict token limit (\texttt{<= 2000 tokens}) and forbids excessive brainstorming, preventing the model from entering infinite reasoning loops and saving computational overhead.
    \item \textbf{Robust Extraction Standard:} Adopting a common standard from mathematical reasoning tasks in RL, the final decision must be wrapped in \texttt{\textbackslash boxed\{\}}. This guarantees deterministic extraction of the predicted player ID during training or evaluation.
\end{itemize}

\newtcblisting{decisionbox}[1][]{
    colback=gray!5,           
    colframe=gray!40!black,   
    listing only,             
    breakable,                
    title=Detection Stage Prompt, 
    fonttitle=\bfseries,      
    sharp corners,            
    boxrule=0.8pt,            
    listing options={
        basicstyle=\ttfamily\small, 
        breaklines=true,            
        columns=fullflexible,
        keepspaces=true             
    },
    #1
}

\begin{decisionbox}
decision_context = f"""You are analyzing a game of "Who's the Odd One Out?" where {num_players} players have just written short stories based on a writing prompt. Your task is to identify which player is the spy (the one who received NO writing prompt) based on their story.

Game Setup:
- There are {num_players} players in total. Only one player (the spy) received NO writing prompt, while all other players received the SAME writing prompt.
- The spy sees nothing (blank prompt).
- Each player wrote a short story trying to match the prompt while avoiding suspicion.
- The spy tries to blend in by inferring the theme from others and writing something that seems consistent.
- Non-spy players try to write stories that are thematically aligned, meaningful, coherent, and creative.

Your task:
Analyze the writing prompt and all the stories and determine which player is most likely the spy. Your primary strategy is to judge who likely had access to the prompt.
Look for:
1. **Off-theme / mismatch**: The story does not match the prompt's theme, constraints, or implied setting.
2. **Shallow / not meaningful**: Lacks subtext, turning point, or deeper implication.
3. **Weak narrative craft**: Straight-line narration, poor coherence, no payoff, abrupt transitions.
4. **Low creativity**: Generic style, cliche patterns, unimaginative choices.
5. **Consistency with others**: Non-spy players tend to converge on the same prompt-driven theme; a spy may drift or imitate superficially.

Efficiency constraints (IMPORTANT):
- Do a fast check for each story (on-theme? meaningful? coherent? creative?) and pick the single most suspicious player.
- If you cannot immediately determine who the mole is, answer \\boxed{{N/A}} first and then continue thinking.  
- Do NOT brainstorm. The entire output must be <= 2000 tokens.

The writing prompt below is what non-spy players see. Compare each story against this reference:

[Reference Writing Prompt]
{writing_prompt}

Required output format:
First, conduct your private reasoning -- may include suspicions, probabilities, evidence analysis, etc.
Then, put your final answer (PLAYER_NUMBER or N/A) inside \\boxed{{}}. If you are uncertain, you can answer N/A.
Example answer: \\boxed{{1}}; \\boxed{{2}}; \\boxed{{3}}; \\boxed{{N/A}}. 
Hard limit: The entire output must be <= 2000 tokens.

All Stories from the Story-writing Stage:
{all_stories}"""
\end{decisionbox}

\begin{table*}[t]
\centering
\setlength{\tabcolsep}{8pt}
\caption{\textbf{A/B test evaluation against the untrained base model on summarization
benchmarks.} Each row reports the GPT-4o win rate (\%) of a trained model against its own
untrained backbone, so $50\%$ denotes parity and the untrained-model rows act as a calibration
reference. \textbf{Avg.} is the mean over the five benchmarks. Shaded rows are SpyRL.}
\label{tab:summarization_vs_base}
\vspace{-5pt}
\resizebox{\textwidth}{!}{
\begin{tabular}{lcccccc}
\toprule[1.5pt]
\textbf{Model}
& \textbf{GovReport} & \textbf{Multi\_News} & \textbf{QmSum}
& \textbf{VcSum} & \textbf{SamSum} & \textbf{Avg.} \\
\midrule
Qwen3-4B          & 51.2\% & 52.1\% & 52.4\% & 51.8\% & 50.9\% & 51.7\% \\
+ R-Zero          & 56.8\% & 48.2\% & 55.2\% & 51.2\% & 48.1\% & 51.9\% \\
+ Absolute Zero   & 58.8\% & 61.2\% & 56.4\% & 62.8\% & 68.5\% & 61.5\% \\
\rule{0pt}{8pt}
\cellcolor{OursRow}\ours{+ SpyRL}
& \cellcolor{OursRow}\ours{74.6\%} & \cellcolor{OursRow}\ours{80.2\%}
& \cellcolor{OursRow}\ours{68.4\%} & \cellcolor{OursRow}\ours{70.2\%}
& \cellcolor{OursRow}\ours{76.2\%} & \cellcolor{OursRow}\ours{73.9\%} \\
\midrule
Qwen3-8B          & 50.2\% & 51.4\% & 52.8\% & 53.1\% & 51.6\% & 51.8\% \\
+ R-Zero          & 51.3\% & 50.2\% & 47.9\% & 55.3\% & 52.9\% & 51.5\% \\
+ Absolute Zero   & 62.5\% & 50.3\% & 53.2\% & 58.2\% & 70.6\% & 59.0\% \\
\rule{0pt}{8pt}
\cellcolor{OursRow}\ours{+ SpyRL}
& \cellcolor{OursRow}\ours{78.2\%} & \cellcolor{OursRow}\ours{68.5\%}
& \cellcolor{OursRow}\ours{78.2\%} & \cellcolor{OursRow}\ours{72.5\%}
& \cellcolor{OursRow}\ours{79.5\%} & \cellcolor{OursRow}\ours{75.4\%} \\
\bottomrule[1.5pt]
\end{tabular}}
\vspace{8pt}
\end{table*}

\begin{table*}[t]
\vspace{-5pt}
\centering
\setlength{\tabcolsep}{4.5pt}
\caption{\textbf{A/B test evaluation against the untrained base model on writing benchmarks.}
Each row reports the GPT-4o win rate of a trained model against its own untrained backbone,
so $50\%$ denotes parity. Higher is better.}
\vspace{-10pt}
\label{tab:writing_vs_base}
\resizebox{\textwidth}{!}{
\begin{tabular}{lccccc ccccc}
\toprule[1.5pt]
& \multicolumn{5}{c}{\textbf{WritingPrompt}} & \multicolumn{5}{c}{\textbf{WritingBench}} \\
\cmidrule(lr){2-6} \cmidrule(lr){7-11}
Model
& \shortstack{Novel}
& \shortstack{Emotion}
& \shortstack{Coher.}
& \shortstack{Consist.}
& \shortstack{Overall}
& \shortstack{Novel}
& \shortstack{Emotion}
& \shortstack{Coher.}
& \shortstack{Consist.}
& \shortstack{Overall} \\
\midrule
Qwen3-4B
& 51.2 & 50.0 & 52.3 & 51.0 & 51.2
& 50.8 & 51.5 & 50.9 & 52.1 & 51.0 \\
+ R-Zero
& 48.3 & 44.3 & 51.2 & 48.8 & 48.8
& 46.5 & 46.5 & 43.2 & 46.2 & 46.5 \\
+ Absolute Zero
& 54.5 & 52.2 & 50.2 & 52.8 & 54.0
& 55.2 & 54.8 & 55.7 & 55.4 & 55.2 \\
\rule{0pt}{8pt}
\cellcolor{OursRow}\textbf{\ours{+ SpyRL}}
& \cellcolor{OursRow}\ours{84.3} & \cellcolor{OursRow}\ours{76.8} & \cellcolor{OursRow}\ours{72.3} & \cellcolor{OursRow}\ours{70.1} & \cellcolor{OursRow}\ours{81.3}
& \cellcolor{OursRow}\ours{76.2} & \cellcolor{OursRow}\ours{75.7} & \cellcolor{OursRow}\ours{68.5} & \cellcolor{OursRow}\ours{68.0} & \cellcolor{OursRow}\ours{75.1} \\
\midrule
Qwen3-8B
& 52.2 & 51.8 & 50.6 & 51.0 & 51.5
& 50.4 & 51.1 & 51.3 & 52.4 & 51.8 \\
+ R-Zero
& 52.3 & 54.2 & 51.2 & 49.5 & 52.2
& 52.3 & 52.1 & 52.5 & 53.4 & 52.0 \\
+ Absolute Zero
& 55.3 & 52.8 & 57.4 & 56.8 & 56.4
& 56.5 & 55.8 & 58.2 & 57.9 & 58.1 \\
\rule{0pt}{8pt}
\cellcolor{OursRow}\textbf{\ours{+ SpyRL}}
& \cellcolor{OursRow}\ours{77.3} & \cellcolor{OursRow}\ours{76.2} & \cellcolor{OursRow}\ours{74.2} & \cellcolor{OursRow}\ours{75.0} & \cellcolor{OursRow}\ours{76.5}
& \cellcolor{OursRow}\ours{78.1} & \cellcolor{OursRow}\ours{77.4} & \cellcolor{OursRow}\ours{71.0} & \cellcolor{OursRow}\ours{71.2} & \cellcolor{OursRow}\ours{78.1} \\
\bottomrule[1.5pt]
\end{tabular}
}
\vspace{3pt}
\end{table*}

\section{Additional Experiments}\label{app:additional}

This section presents additional experiments that complement the main results from different perspectives. These experiments further examine the effectiveness, training stability, cross-task transferability, evaluation robustness, and human alignment of SpyRL, providing a more comprehensive validation of the proposed framework.

\subsection{A/B Evaluation against the Untrained Base Model}
\label{app:vs_base}

The A/B evaluations in the main text are anchored on the competing method: each row of
Tables~\ref{tab:summarization_results} and~\ref{tab:additional_abtest_writing} compares SpyRL
directly against one opponent. That presentation answers which method a reader should prefer, but
it does not show how much each \emph{baseline} gains from its own self-evolution procedure. For
completeness, Tables~\ref{tab:summarization_vs_base} and~\ref{tab:writing_vs_base} therefore report
the complementary view for summarization and creative writing, in which every method is compared
against its \emph{own} untrained backbone under the same swapped-order protocol. A win rate of
$50\%$ denotes parity, and the untrained-model rows act as a calibration reference.

Three observations follow. First, the untrained backbones score close to chance against themselves,
with average win rates of $51.7\%$ and $51.8\%$ on summarization and overall win rates between
$51.0\%$ and $51.8\%$ on creative writing. This confirms that the swapped-order aggregation
described in Appendix~\ref{app:baselines} removes most of the position bias, so deviations from
$50\%$ elsewhere in the table reflect genuine quality differences.

Second, the two self-evolution baselines move the needle very little on either task. On
summarization, R-Zero averages $51.9\%$ and $51.5\%$ for Qwen3-4B and Qwen3-8B, i.e.\ statistically
indistinguishable from no training at all, while Absolute Zero reaches $61.5\%$ and $59.0\%$. On
creative writing the gap is starker: Absolute Zero attains overall win rates of only $54.0\%$ and
$55.2\%$ with Qwen3-4B and $56.4\%$ and $58.1\%$ with Qwen3-8B, and R-Zero is preferred \emph{less}
often than the untrained Qwen3-4B ($48.8\%$ and $46.5\%$), meaning that its training actively
degrades creative writing. SpyRL, by contrast, averages $73.9\%$ and $75.4\%$ on summarization and
attains overall win rates of $81.3\%$ and $75.1\%$ with Qwen3-4B and $76.5\%$ and $78.1\%$ with
Qwen3-8B on creative writing.

Third, this contrast follows directly from how the three methods obtain their reward. R-Zero and
Absolute Zero require verifiable solver feedback to calibrate task difficulty, which is unavailable
for open-ended generation, so their proposer--solver loop has nothing reliable to optimize against.
SpyRL instead derives its reward from the environment-assigned spy identity, which remains well
defined regardless of whether the target task admits a verifier. Together with the main-text
results, this confirms that the gains of SpyRL are not an artifact of the choice of evaluation
anchor.

\subsection{Evaluation on More Challenging Reasoning Benchmarks}
\label{app:harder_reasoning}

The reasoning benchmarks in the main experiments span a wide difficulty range, but the hardest of them (AIME 2024 and AIME 2025) contain only 30 problems each, so we additionally evaluate on three larger and more demanding benchmarks: AMC, Olympiad-Bench, and SuperGPQA. We reuse the Qwen3-4B checkpoints trained under our mathematical reasoning setting without any further fine-tuning, and compare Qwen3-4B, R-Zero, Absolute Zero, and SpyRL. We clarify that the Absolute Zero baseline in this work is not the code-oriented checkpoint released with the original paper. Instead, we reimplement and train Absolute Zero with the same Qwen3-4B backbone under our math-domain setting, ensuring a controlled and fair comparison across methods.

As shown in Table~\ref{tab:reasoning_benchmarks}, SpyRL achieves the best performance on all three benchmarks, reaching 55.0 on AMC, 42.7 on Olympiad-Bench, and 30.1 on SuperGPQA, compared with 47.5, 34.8, and 25.4 for the untrained backbone. Averaged over the three benchmarks this is a gain of 6.7 points, against 1.6 for R-Zero and 3.6 for Absolute Zero. SpyRL is also the only method that improves on every benchmark: R-Zero falls below the untrained backbone on AMC, indicating that its proposer--solver curriculum does not transfer to competition-style problems.

These results demonstrate that the improvements of SpyRL extend beyond the benchmarks reported in the main text and remain effective on more difficult reasoning problems. We attribute this advantage to the information-asymmetric self-play mechanism: incomplete derivations, overlooked conditions, and unsupported assumptions are more likely to expose the spy player during detection, allowing these reasoning deficiencies to be converted into verifiable training signals. Consequently, SpyRL encourages the model to produce more complete, rigorous, and reliable reasoning processes.

\begin{table*}[t]
\vspace{-10pt}
\centering
\setlength{\tabcolsep}{8pt}
\caption{\textbf{Results on harder reasoning benchmarks}. We report accuracy (\%) of Qwen3-4B and
its self-evolution variants on AMC, Olympiad-Bench, and SuperGPQA, using the same math-domain
checkpoints as Table~\ref{tab:main_results}. \textbf{Avg.} is the mean over the three benchmarks
and \textbf{$\Delta$} the gain over the untrained backbone. Absolute Zero is reimplemented on the
same backbone under our math-domain setting rather than taken from the released code-oriented
checkpoint, so that all methods are trained comparably. The shaded row is SpyRL.}
\label{tab:reasoning_benchmarks}
\vspace{-5pt}
\resizebox{\textwidth}{!}{
\begin{tabular}{lccccc}
\toprule[1.5pt]
\textbf{Method} & \textbf{AMC} & \textbf{Olympiad-Bench} & \textbf{SuperGPQA}
& \textbf{Avg.} & \textbf{$\Delta$ vs. Base} \\
\midrule
Qwen3-4B          & 47.5 & 34.8 & 25.4 & 35.9 & --  \\
+ R-Zero          & 45.0 & 39.7 & 27.8 & 37.5 & $+1.6$ \\
+ Absolute Zero   & 50.0 & 41.5 & 27.1 & 39.5 & $+3.6$ \\
\rule{0pt}{8pt}
\cellcolor{OursRow}\ours{+ SpyRL}
& \cellcolor{OursRow}\ours{55.0}
& \cellcolor{OursRow}\ours{42.7}
& \cellcolor{OursRow}\ours{30.1}
& \cellcolor{OursRow}\ours{42.6}
& \cellcolor{OursRow}\ours{$+6.7$} \\
\bottomrule[1.5pt]
\end{tabular}}
\vspace{8pt}
\end{table*}

\subsection{Joint versus Alternating Two-Stage Optimization}
\label{app:joint_vs_alternating}

The performing and detection stages of SpyRL are mutually dependent: the detector's votes determine the rewards used to optimize the performer, while the outputs generated by the performer constitute the training inputs for the detector. A straightforward implementation is to jointly update both stages within every training epoch. However, this strategy introduces a tightly coupled and highly non-stationary optimization process, particularly during the early stages of training. At initialization, the detector has limited ability to identify the spy player, while both civilian and spy players produce relatively weak task outputs. Consequently, the detector's votes may not reliably reflect differences in performing quality, resulting in noisy or misleading rewards for the performing stage. Updating the performer with such inaccurate rewards can further degrade its outputs, which in turn provides the detector with less informative training examples. 

To alleviate this issue, SpyRL adopts an alternating optimization strategy in which only one stage is updated during each training epoch while the other stage remains fixed. In particular, strengthening the detector before using its voting outcomes to optimize the performer provides a more reliable reward signal for the performing stage. Once the detector can meaningfully distinguish outputs generated under complete and degraded information, the performer is encouraged to produce more rigorous and strategically convincing responses. These improved outputs subsequently increase the difficulty of identity detection and provide more informative examples for further detector training. Alternating optimization therefore decomposes the coupled learning problem into more stable stage-wise updates and allows the performing and detection capabilities to progressively shape each other.

The resulting performance on five mathematical reasoning benchmarks is reported in Table~\ref{tab:joint_alternating}. Joint training fails to improve the base model and substantially degrades performance on several benchmarks. For example, accuracy decreases from 84.5 to 76.8 on GSM8K, from 68.2 to 53.1 on Math500, and from 42.3 to 33.1 on Minerva. Its average score across the five benchmarks drops from 42.4 to 35.3.

In contrast, alternating optimization consistently improves the model across all evaluated benchmarks. It increases GSM8K accuracy from 84.5 to 93.4 and Math500 accuracy from 68.2 to 79.5. The improvement is particularly pronounced on AIME 2025, where accuracy rises from 6.7 to 20.0. Averaged across the five benchmarks, alternating training achieves a score of 50.8, corresponding to an improvement of 8.4 points over the base model and 15.5 points over joint training. These results support our analysis that simultaneously updating the two mutually dependent stages leads to inaccurate credit assignment and unstable learning, whereas alternating optimization provides sufficiently reliable intermediate signals for sustained co-evolution between the performer and detector.





\begin{table*}[t]
\centering
\caption{\textbf{Comparison of joint and alternating two-stage optimization.}
We report the accuracy of Qwen3-4B. Joint training updates the performer and detector simultaneously, whereas alternating training updates only one stage in each epoch.}
\label{tab:joint_alternating}
\vspace{-5pt}
\resizebox{0.7\textwidth}{!}{
\begin{tabular}{cccccc}
\toprule[1.5pt]
  & \textbf{GSM8K} & \textbf{Math500} & \textbf{AIME 24} & \textbf{AIME 25} & \textbf{Minerva} \\
\midrule
\midrule
\textbf{Qwen3-4b} & 84.5 & 68.2 & 10.3 & 6.7 & 42.3 \\
\midrule
\textbf{+ Joint Training} & 76.8 & 53.1 & 6.7 & 6.7 & 33.1 \\
\midrule
\textbf{+ Alternative Training} & 93.4 & 79.5 & 13.3 & 20.0 & 47.8 \\
\bottomrule[1.5pt]
\end{tabular}}
\vspace{10pt}
\end{table*}

\subsection{Evaluation with an Alternative LLM Judge}
\label{app:gemini_evaluation}

To examine whether the improvements of SpyRL are robust to the choice of automatic evaluator, we repeat the A/B evaluations on summarization and creative writing using Gemini-3.5-Flash as an alternative judge. We follow the same evaluation protocol as in the main experiments: for each test instance, the evaluator compares the anonymized outputs of the SpyRL-trained model and its corresponding original Qwen3 model. We also aggregate judgments obtained under swapped response orders to mitigate position bias. The reported value is the proportion of comparisons in which the evaluated model is preferred over the original model.

As shown in Table~\ref{tab:gemini_summary}, SpyRL consistently improves summarization performance across all five benchmarks. The Qwen3-4B model trained with SpyRL obtains an average win rate of $79.8\%$, compared with $52.3\%$ for the original model, while the corresponding Qwen3-8B model achieves an average win rate of $76.1\%$, compared with $52.0\%$ for its base model. The improvements are consistent across GovReport, Multi-News, QMSum, VCSum, and SAMSum, indicating that the gains are not limited to a particular summarization domain.

A similar trend is observed for creative writing in Table~\ref{tab:gemini_writing}. SpyRL substantially outperforms the original models across novelty, emotion, coherence, consistency, and overall quality on both WritingPrompts and WritingBench. In particular, Qwen3-4B achieves overall win rates of $81.8\%$ and $81.4\%$ on the two benchmarks, while Qwen3-8B achieves $77.8\%$ and $77.0\%$, respectively. These results are consistent with the GPT-4o-based evaluation in the main experiments, demonstrating that the observed improvements remain stable under a different LLM judge and are unlikely to arise from evaluator-specific preferences.

\begin{table*}[t]
\centering
\caption{\textbf{Gemini-3.5-Flash A/B evaluation on summarization benchmarks.}
We report the win rates of the original Qwen3 models and their SpyRL-trained counterparts against the corresponding base model across five summarization benchmarks. Higher is better.}
\label{tab:gemini_summary}
\vspace{-5pt}
\resizebox{0.7\textwidth}{!}{
\begin{tabular}{cccccc}
\toprule[1.5pt]
 & \textbf{GovReport} &\textbf{Multi\_News}  & \textbf{QmSum} & \textbf{VcSum}  & \textbf{SamSum} \\
\midrule
\midrule
\textbf{Qwen3-4b} & 52.4\% & 51.9\% & 53.8\% & 52.0\% & 51.6\% \\
\midrule
\textbf{+SpyRL} & 81.3\% & 85.2\% & 74.9\% & 78.8\% & 79.0\% \\
\midrule
\midrule
\textbf{Qwen3-8b} & 50.6\% & 52.3\% & 51.8\% & 52.2\% & 53.1\% \\
\midrule
\textbf{+SpyRL} & 79.3\% & 70.6\% & 76.4\% & 75.9\% & 78.2\% \\
\bottomrule[1.5pt]
\end{tabular}}
\vspace{8pt}
\end{table*}

\begin{table*}[t]
\vspace{-5pt}
\centering
\setlength{\tabcolsep}{4.5pt}
\caption{\textbf{Gemini-3.5-Flash A/B evaluation on creative writing benchmarks.}
We report win rates across five level quality on WritingPrompts and WritingBench.}
\label{tab:gemini_writing}
\vspace{-10pt}
\resizebox{\textwidth}{!}{
\begin{tabular}{lccccc ccccc}
\toprule[1.5pt]
& \multicolumn{5}{c}{\textbf{WritingPrompt}} & \multicolumn{5}{c}{\textbf{WritingBench}} \\
\cmidrule(lr){2-6} \cmidrule(lr){7-11}
& \shortstack{Novel}
& \shortstack{Emotion}
& \shortstack{Coher.}
& \shortstack{Consist.}
& \shortstack{Overall}
& \shortstack{Novel}
& \shortstack{Emotion}
& \shortstack{Coher.}
& \shortstack{Consist.}
& \shortstack{Overall} \\
\midrule
\midrule
\textbf{Qwen3-4B} 
& 52.5\% & 51.4\% & 52.9\% & 53.1\% & 52.5\%
& 53.4\% & 52.7\% & 53.2\% & 51.4\% & 52.7\% \\
\midrule
\textbf{+ SpyRL} 
& 88.7\% & 82.5\% & 79.1\% & 76.9\% & 81.8\%
& 85.7\% & 84.9\% & 78.3\% & 76.5\% & 81.4\% \\
\midrule
\textbf{Qwen3-8B} 
& 52.9\% & 53.1\% & 50.2\% & 52.8\% & 52.3\%
& 50.3\% & 52.7\% & 52.9\% & 53.6\% & 52.4\% \\
\midrule
\textbf{+ SpyRL} 
& 78.1\% & 80.7\% & 75.0\% & 77.3\% & 77.8\%
& 79.4\% & 78.1\% & 74.6\% & 75.8\% & 77.0\% \\
\bottomrule[1.5pt]
\end{tabular}
}
\vspace{3pt}
\end{table*}

\subsection{Agreement between GPT-4o and Human Evaluation}
\label{app:gpt_human_agreement}

To examine the reliability of GPT-4o as an automatic evaluator, we measure its agreement with the human annotations collected in our creative-writing evaluation. Treating the human judgments as ground-truth labels, we formulate the pairwise preference for SpyRL as a binary classification problem and compute the precision and recall of GPT-4o for each evaluation dimension. Here, precision measures the proportion of samples judged as wins for SpyRL by GPT-4o that are also preferred by human evaluators, while recall measures the proportion of human-preferred SpyRL samples that are correctly identified by GPT-4o.

As shown in Table~\ref{tab:gpt_human_agreement}, GPT-4o exhibits strong agreement with human judgments across all five dimensions. Precision ranges from $85.7\%$ to $91.0\%$, while recall ranges from $79.4\%$ to $93.8\%$. In particular, GPT-4o achieves strong performance on novelty, emotion, and overall quality, with an overall precision of $91.0\%$ and recall of $93.8\%$. Although the recall for consistency is relatively lower at $79.4\%$, the results remain consistently high overall. These findings indicate that GPT-4o provides a reliable approximation of human preferences and support its use as an automatic evaluator for comparing the quality of open-ended creative-writing outputs.

\begin{table*}[h]
\vspace{-5pt}
\centering
\setlength{\tabcolsep}{4.5pt}
\caption{\textbf{Agreement between GPT-4o and human evaluation.}
Treating human judgments as ground-truth labels, we report the precision and recall of GPT-4o in identifying pairwise wins for SpyRL across five creative-writing evaluation dimensions. Higher is better.}
\label{tab:gpt_human_agreement}
\vspace{-10pt}
\resizebox{0.7\textwidth}{!}{
\begin{tabular}{lccccc}
\toprule[1.5pt]
& \shortstack{Novel}
& \shortstack{Emotion}
& \shortstack{Coher.}
& \shortstack{Consist.}
& \shortstack{Overall} \\
\midrule
\midrule
Precision & 90.1\% & 85.7\% & 89.1\% & 86.7\% & 91.0\% \\
\midrule
Recall & 89.7\% & 89.4\% & 87.2\% & 79.4\% & 93.8\% \\
\bottomrule[1.5pt]
\end{tabular}
}
\vspace{3pt}
\end{table*}

\end{document}